\documentclass[10pt,twocolumn,letterpaper]{article}

\PassOptionsToPackage{table}{xcolor}
\usepackage[pagenumbers]{cvpr}

\usepackage{algorithm}
\usepackage[noend]{algpseudocode}
\usepackage{soul}

\algnewcommand{\IfThen}[2]{\State \algorithmicif\ #1\ \algorithmicthen\ #2}
\AtEndPreamble{\crefname{algorithm}{Alg.}{Algs.}\Crefname{algorithm}{Algorithm}{Algorithms}}

\usepackage{amsthm}

\usepackage{booktabs}
\usepackage[table]{xcolor}
\usepackage{graphicx}
\definecolor{memrowblue}{RGB}{221,236,251}
\definecolor{memsectiongray}{RGB}{240,242,245}

\AtEndPreamble{%
  \theoremstyle{plain}%
  \newtheorem{theorem}{Theorem}%
  \newtheorem{proposition}[theorem]{Proposition}%
  \theoremstyle{definition}%
  \theoremstyle{remark}%
  \crefname{assumption}{Assumption}{Assumptions}%
}

\definecolor{cvprblue}{rgb}{0.21,0.49,0.74}
\usepackage[pagebackref,breaklinks,colorlinks,allcolors=cvprblue]{hyperref}

\usepackage{pgfplots}
\usepgfplotslibrary{groupplots}

\pgfplotsset{compat=1.18}

\title{Memryoshka: Adaptive-Fidelity Video Memory for Streaming VLMs}

\author{%
Ahmadreza Jeddi$^{1,2,3*\dagger}$ \quad
Enming Zhang$^{1,2*}$ \quad
Jasper Gerigk$^{1,2}$ \quad
Shivam Chandhok$^{2,4}$ \\
Mozhgan Nasr Azadani$^{5,6}$ \quad
Hakki Can Karaimer$^{7}$ \quad
Yize Liu$^{5}$ \quad
Leonid Sigal$^{2,4}$ \\
Igor Gilitschenski$^{1,2}$ \quad
Konstantinos G. Derpanis$^{1,2,8}$ \quad
Marco Pavone$^{5,9}$ \quad
Babak Taati$^{1,2,3}$ \\[8pt]
{\normalsize
$^1$University of Toronto \quad
$^2$Vector Institute \quad
$^3$University Health Network \quad
$^4$University of British Columbia} \\
{\normalsize
$^5$Stanford University \quad
$^6$University of Waterloo \quad
$^7$Samsung AI Center Toronto \quad
$^8$York University \quad
$^9$NVIDIA} \\[4pt]
{\small $^*$Equal contribution \quad $^\dagger$Corresponding author: \href{mailto:ajeddi@cs.toronto.edu}{\texttt{ajeddi@cs.toronto.edu}}}
}

\begin{document}
\maketitle
\begin{abstract}
Existing streaming VLM memory systems manage limited
storage by selecting, merging, or discarding visual
evidence, often retaining representations at fixed
precision.
We observe that visual tokens are highly compressible along
their feature dimension with little loss in downstream accuracy.
This reveals a key trade-off: under a fixed memory budget,
preserving broader temporal coverage at lower fidelity may
be preferable to retaining fewer observations at high precision.
We introduce Memryoshka, a training-free adaptive-fidelity
memory that jointly allocates storage across frames, tokens,
and precision.
Visual tokens are stored as nested bitstreams, where
longer prefixes recover increasingly finer detail.
A shared price per bit determines which evidence to retain
and at what precision, allowing stored codes to be progressively
shortened without revisiting past frames.
Across long-video and streaming benchmarks, Memryoshka matches
the average accuracy of an uncapped full-precision memory under
the same prefill context budget, using just 7.2\,MB of storage,
roughly 1,000 times less than storing one hour
of full-precision visual tokens.
It substantially improves performance over existing
visual-memory baselines.
These results highlight the importance of jointly adapting
temporal coverage and representation fidelity for efficient
long-term visual memory.

\vspace{6pt}
\centerline{\upshape\bfseries Project page:\enspace
\href{https://armenjeddi.github.io/memryoshka/}{\textcolor[HTML]{B83228}{armenjeddi.github.io/memryoshka}}}
\end{abstract}

\section{Introduction}
\label{sec:intro}

\begin{figure}[t]
  \centering

  \begin{subfigure}[t]{\linewidth}
    \centering
    \includegraphics[width=\linewidth]{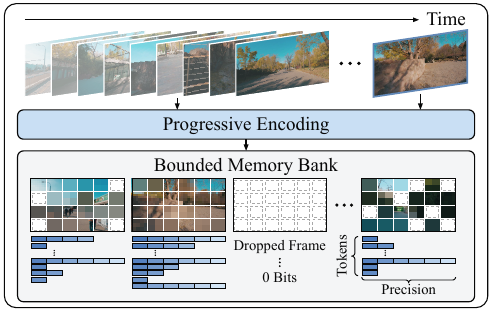}
    \caption{Adaptive Memory Bank}
    \label{fig:teaser_a}
  \end{subfigure}

  \begin{subfigure}[t]{0.49\linewidth}
    \centering
    \includegraphics[width=\linewidth]{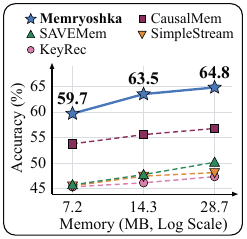}
    \caption{Overall Accuracy}
    \label{fig:teaser_b}
  \end{subfigure}
  \hfill
  \begin{subfigure}[t]{0.49\linewidth}
    \centering
    \includegraphics[width=\linewidth]{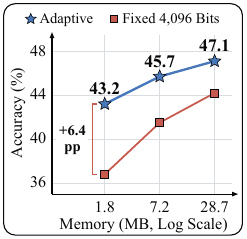}
    \caption{Adaptive Precision}
    \label{fig:teaser_c}
  \end{subfigure}

  \vspace{-0.5em}

  \caption{\textbf{Memryoshka: Adaptive-fidelity visual memory.}
  (a) As video streams grow, Memryoshka uses a shared price per bit to jointly allocate a fixed memory budget across frames, tokens, and precision.
  (b) Memryoshka outperforms existing visual-memory baselines across storage budgets.
  (c) Compared with fixed-precision PCA coding (512 components, 8 bits each), adaptive fidelity consistently improves performance.}

  \vspace{-1em}
  \label{fig:teaser}
\end{figure}

\begin{quote}
\emph{``The pictures drawn in our minds are laid in fading colours;
and if not sometimes refreshed, vanish and disappear.''}

\hfill --- John Locke, \emph{An Essay Concerning Human Understanding} (1690)
\end{quote}

Streaming vision-language models must answer questions about continuously growing
video histories, as in wearable assistants, robots, and real-time monitoring
systems~\cite{chen2024videollmonline,qian2025dispider,yan2025teleego}.
Unlike conventional long-video question answering, where the video and question
are typically available together, streaming systems must decide what visual
evidence to preserve as frames arrive, without knowing what future questions
may require.
This creates a central challenge: \emph{retaining potentially useful information
over long time horizons under a fixed memory budget.}

Storing all visual tokens quickly becomes impractical as video streams grow.
For example, sampling one $448\times448$ frame per second with
Qwen2.5-VL-7B~\cite{bai2025qwen25vl} produces approximately
920{,}000 visual tokens per hour, requiring 6.6\,GB of storage
and exceeding the model's context window by more than sixfold.
A memory system can reduce this stream at three granularities: which frames to keep,
which tokens within each frame, and how many bits per token.
Existing streaming memories focus mainly on the first two.
Many select, merge, or evict visual tokens and pass the resulting
bounded memory directly to the LLM
reader~\cite{zhang2024flashvstream,song2024moviechat,kim2025infinipotv,yang2025streammem,song2026causalmem},
tying storage to the reader's input budget.
Retrieval-based memories instead separate storage from readout,
preserving more evidence than the reader can process at once
and selecting relevant information when questions
arrive~\cite{qian2024videostreaming,di2025rekv,wu2026savemem,chen2026keyrec}.
These approaches primarily determine \emph{which} evidence to retain,
often at fixed precision even when quantization is used~\cite{chen2026streamingtom}.
Yet visual tokens are highly compressible along their feature dimension:
we find that even simple quantized PCA representations substantially
reduce storage with little loss in downstream accuracy.
This suggests that, under a fixed budget, preserving more evidence
at lower fidelity may be preferable to retaining fewer
full-precision tokens.
Since every bit spent on precision is a bit not spent on another
frame or token, this raises the question of \emph{how to allocate
a fixed memory budget across frames, tokens, and precisions as
the stream grows.}

We introduce \textbf{Memryoshka} (Memory Matryoshka),
a training-free adaptive-fidelity memory for streaming VLMs
that allocates a fixed storage budget across frames, tokens,
and their precision using a shared price per bit.
Visual tokens are encoded as nested bitstreams, whose
prefixes provide progressively finer reconstructions
and can be shortened without re-encoding (\cref{fig:teaser_a}).
An online writer estimates frame novelty, token importance,
and the value of each code length to decide what to retain
and at what precision.
As the stream grows, the shared price rises, allowing
stored codes to lose precision or be discarded without
revisiting past frames.
When a question arrives, a lightweight query-conditioned
readout retrieves and decodes relevant evidence for the
frozen VLM reader under a separate readout budget.

We evaluate Memryoshka on six long-video and streaming benchmarks
with a frozen Qwen2.5-VL-7B reader.
Across memory budgets, Memryoshka improves average accuracy
by 6--8 percentage points over existing visual-memory baselines
(\cref{fig:teaser_b}), while matching the performance of uncapped
full-precision memory at a fraction of its storage cost.
Our analysis further shows that adaptive allocation across
frames, tokens, and representation fidelity matches or
outperforms the best fixed-precision configurations across
nearly all settings (\cref{fig:teaser_c}).
These results demonstrate the importance of adaptive fidelity
for efficient long-term visual memory. Our contributions are threefold:
\begin{enumerate}
    \item \textbf{Joint allocation across frames, tokens, and precisions.}
    We formulate streaming visual memory as a joint allocation problem,
    studying the trade-off between temporal coverage and representation fidelity.

    \item \textbf{Training-free adaptive-fidelity memory.}
    We introduce nested token codes and a shared price per bit
    that adapts memory allocation as the stream grows, without training.

    \item \textbf{Effectiveness across tasks and modalities.}
    Memryoshka substantially outperforms visual-memory baselines
    across six benchmarks, approaches uncapped-memory performance
    with far less storage, and extends to robot policies and
    audio-visual streams.
\end{enumerate}

\section{Related Work}
\label{sec:related}

\paragraph{Long-video and streaming understanding.}
Offline long-video benchmarks~\cite{fu2024videomme,wang2024lvbench,wu2024longvideobench,zhou2024mlvu}
provide access to the full video and question, allowing methods
to select relevant frames~\cite{liu2025bolt,sun2025mdp3,tang2025aks,wang2024videoagent,yu2024framevoyager,zhang2025qframe}
or reduce visual-token costs through pruning or
merging~\cite{bolya2023tome,chen2024fastv,fu2024framefusion,li2023llamavid,shang2024prumerge,shen2024longvu,tao2024dycoke}.
Related efficiency methods prune visual tokens without training~\cite{jeddi2025saint}, adapt how much visual context and reasoning each input receives~\cite{jeddi2026avis}, or distill full-context behavior into models that read pruned tokens~\cite{jeddi2026scopd}.
In contrast, streaming video understanding requires processing
frames as they arrive, often before future questions are
known~\cite{huang2025ovbench,li2025ovobench,lin2024streamingbench,yan2025teleego}.
Existing approaches study efficient online frame processing
and temporal understanding~\cite{chen2024videollmonline,qian2025dispider,wang2025streambridge,wu2024videollmmod,xu2026streamingvlm}.
However, recent studies show that retaining only a few recent
frames can perform competitively on streaming
benchmarks~\cite{shen2026simplestream,zhang2026streamarena},
suggesting that high scores do not necessarily reflect
long-term recall.
Memryoshka addresses question-agnostic memory construction,
preserving evidence over long video streams under a fixed
storage budget through adaptive fidelity.
\vspace{-0.5em}

\paragraph{Streaming visual memories.}
Existing methods manage limited storage by merging
frames~\cite{he2024malmm,song2024moviechat,wang2025videollamb},
preserving different levels of detail~\cite{li2026freshmem,zeng2025streamforest,zhang2024flashvstream},
or selecting and evicting visual tokens~\cite{chen2026streamingtom,wang2025stc,xie2026fluxmem,yao2025timechatonline}.
Others use online selection~\cite{song2026causalmem}
or recurrent updates~\cite{zhong2026prem}.
KeyRec~\cite{chen2026keyrec} and SAVEMem~\cite{wu2026savemem}
separate storage and retrieval budgets, selecting which
tokens to preserve and retrieve.
Despite these differences, most focus on which evidence
to retain, keeping representations at fixed precision.
Memryoshka instead jointly adapts temporal coverage and
fidelity across frames and tokens using nested codes
and a shared price per bit.

\vspace{-0.5em}

\paragraph{KV-cache memories.}
Language-model memory methods reduce KV-cache growth by
retaining recent or important entries~\cite{xiao2024streamingllm,li2024snapkv,zhang2023h2o},
compressing older entries~\cite{rae2020compressive},
or retrieving cached representations~\cite{tang2024quest,xiao2024infllm}.
Similar approaches extend to video
streams~\cite{di2025rekv,kim2025infinipotv,yang2025streammem}.
However, they typically process incoming visual evidence
through the LLM, incurring substantial computation and memory costs.
Memryoshka instead compresses visual tokens before LLM
processing, enabling lightweight memory writing while
remaining complementary to KV-cache compression.
\vspace{-0.5em}

\paragraph{Mixed precision and nested codes.}
Quantization reduces KV-cache and visual-token
storage~\cite{hooper2024kvquant,liu2024kivi,tao2025vidkv}.
Mixed-precision methods preserve otherwise evicted tokens
at lower precision~\cite{yang2024mikv}, allocate more bits
to important entries~\cite{he2024zipcache}, reduce the
precision of older entries~\cite{dzikanyanga2026ttkv},
or jointly consider quantization and eviction~\cite{zhang2026rdkv}.
Other methods allocate bits across cached representations
or decorrelated components~\cite{li2026breadthkv,hayes2020remind,wang2022mrdc},
while recent work compresses visual features for
device--edge inference~\cite{li2026sage,pang2026qtofc,yuan2025tofc}.
Concurrent work, RDTok~\cite{zhang2026rdtok}, uses query-independent
rate--distortion optimization to allocate bits across visual
tokens and PCA coefficients, retaining all tokens at
individually optimized rates.
Progressive and nested representations support decoding at
different fidelities through hierarchical dimensions~\cite{kusupati2022matryoshka,lin2024matryoshkakv,rippel2014nested},
nested bit widths~\cite{nair2025matryoshkaquant},
token selection~\cite{cai2024m3,hu2024mqt},
or embedded image codes~\cite{lee2022dpict,lu2021plonq,said1996spiht,shapiro1993ezw,taubman2000high,toderici2016variable}.
Classical rate--distortion methods provide a framework for
bit allocation under shared constraints~\cite{everett1963lagrange,shoham1988efficient}.
Memryoshka brings these ideas to bounded streaming VLM memory
using training-free nested codes and a shared price that
jointly determines which frames and tokens to retain and
at what precision.
Stored codes can be progressively shortened as the stream
grows without re-encoding.

\section{Method}
\label{sec:method}

\subsection{Problem Formulation and Overview}
\label{sec:setting}

\paragraph{Problem Setting.}
Consider a video stream $\mathcal{S}=\{x_1,x_2,\ldots\}$,
where each frame $x_t$ arrives at timestamp $\tau_t$.
We study streaming video question answering, where visual
evidence must be stored before future questions are known.
The system maintains a bounded memory under a fixed storage
budget $B$, without access to past frames for re-encoding.
When a question arrives, the LLM reader processes at most $L$
positions drawn from this memory.
We treat the two budgets separately: $B$ bounds
what is kept over time, and $L$ bounds what is read per question.
The memory can spend $B$ at three granularities: which frames to keep, which tokens to keep, and how many bits to spend on each kept token.
The goal is to divide $B$ among them so that as much of the growing video history as
possible remains retrievable, at a fidelity the reader can still use.

\begin{figure*}[t]
    \centering
    \includegraphics[width=\textwidth]{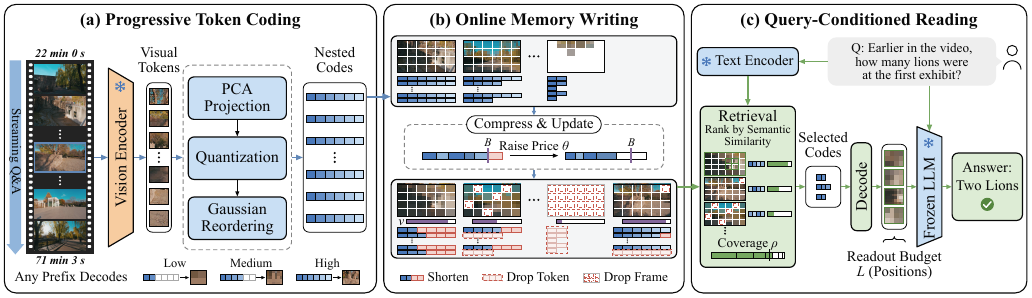}
    \caption{\textbf{Overview of Memryoshka.} (a) Visual tokens from each incoming frame are encoded once into nested bitstreams that support progressive
decoding. (b) An online writer uses a shared price per bit to allocate a fixed storage budget across frames, tokens, and precision, shortening or
discarding stored codes as the stream grows. (c) At question time, query-conditioned retrieval selects and decodes relevant evidence for the frozen
VLM reader under a separate readout budget.}
\vspace{-1em}
    \label{fig:overview}
\end{figure*}

\paragraph{Overall Framework.}
Our framework is illustrated in \cref{fig:overview}.
For each incoming frame $x_t$, the frozen vision encoder
of a VLM produces $N$ visual tokens
$V_t=\{v_{t,i}\}_{i=1}^{N}$, with
$v_{t,i}\in\mathbb{R}^{D}$.
Following the retrieval design of Q-Frame~\cite{zhang2025qframe},
we also use a frozen CLIP-like image encoder~\cite{radford2021clip, zhang2024longclip}
to extract a unit-norm retrieval embedding
$g_t=\phi_{\mathcal{V}}(x_t)$. Memryoshka consists of three stages:

\noindent
\textbf{(I) Progressive Token Coding.}
A training-free nested code stores each visual token as a
progressively decodable bitstream, allowing stored codes
to be shortened without re-encoding
(Sec.~\ref{sec:codec}).

\noindent
\textbf{(II) Online Memory Writing.}
An online writer maintains a bounded memory $\mathcal{M}_t$
by allocating storage across frames, tokens, and their
precision using a shared price per bit:
\begin{equation}
\mathcal{M}_t=\operatorname{Write}
(\mathcal{M}_{t-1},V_t,g_t,\tau_t;B),
\; \; \; \operatorname{bits}(\mathcal{M}_t)\leq B.
\end{equation}
As the stream grows, the writer shortens stored codes or removes them when necessary,
so evidence can lose detail before it becomes unretrievable (Sec.~\ref{sec:write}).

\noindent
\textbf{(III) Query-Conditioned Reading.}
Given a question $q$ at time $t_q$, we use the
CLIP-like text encoder to obtain its embedding
$g_q=\phi_{\mathcal{T}}(q)$, which guides the
retrieval of relevant visual evidence from memory
under the readout budget $L$:
\begin{equation}
\mathcal{C}_q=\operatorname{Read}
(\mathcal{M}_{t_q},g_q;L),
\quad \operatorname{len}(\mathcal{C}_q)\leq L.
\end{equation}
The frozen LLM reader then answers $q$ using the retrieved context $\mathcal{C}_q$
(Sec.~\ref{sec:read}).

\subsection{Progressive Token Coding}
\label{sec:codec}

To adapt memory as the video history grows, we encode
visual tokens into nested bitstreams that can be
progressively shortened.
Our approach is training-free, requiring only a PCA
basis fitted once on unlabeled video, and follows the
principle of embedded coding~\cite{taubman2000high}.

\paragraph{Nested Token Codes.}
Given a visual token $v\in V_t$, we project it onto
a compact PCA basis $W\in\mathbb{R}^{D\times d}$:
\begin{equation}
\label{eq:codec_enc}
z=\bigl(W^\top(v-\mu)\bigr)\oslash\sigma,
\end{equation}
where $\mu$ and $\sigma$ are the PCA mean and component
standard deviations, and $\oslash$ denotes elementwise
division.
We uniformly quantize each component and order the resulting
bits across components by $\sigma_k^2\gamma(p)$, the expected
reduction in squared reconstruction error from bit $p$
of component $k$, where $\gamma(p)$ is the corresponding
reduction for a unit-variance Gaussian under the quantizer.
The ordering preserves bit significance within each component.
Any prefix of $r$ bits can then be decoded to
approximate the original token, with longer prefixes
providing higher fidelity.

\paragraph{Prefix Value.}
Under the independent Gaussian PCA model, assuming diminishing
marginal gains across successive bit planes, this ordering
minimizes expected squared reconstruction error for every
prefix within the given quantization scheme
~\cite{shoham1988efficient,fox1966}.
We define the value of a prefix as the fraction of PCA
variance explained under this model:
\begin{equation}
\label{eq:er}
e(r) = \textstyle\sum_{(k,p)\in P_r}\sigma_k^2\gamma(p) \,/\, \sum_k\sigma_k^2,
\end{equation}
where $P_r$ contains the first $r$ bits in the ordering.
This value is computed directly from the PCA spectrum
and quantizer, without training.
Further derivations and decoding details are provided
in the supplement.

\subsection{Online Writing with a Shared Price}
\label{sec:write}

Following Lagrangian rate--distortion allocation~\cite{everett1963lagrange,shoham1988efficient},
we balance token utility and storage cost using a shared
price $\theta$ per bit. For each incoming frame, the
writer decides which tokens to retain and at what precision.
As memory fills, the same price applies to previously
stored codes, allowing them to lose precision without
re-encoding or be discarded when their utility no longer
justifies their cost.

\paragraph{Price and Decisions.}
Each token can either be discarded ($o=\varnothing$)
or retained with a prefix of length $o=r\in\mathcal{R}$.
We choose the option that maximizes its utility after
accounting for the price of retaining it:
\begin{equation}
\begin{aligned}
o^\star_{t,i}(\theta)
&=\operatorname*{arg\,max}_{o\in\{\varnothing\}\cup\mathcal{R}}
\bigl[u_{t,i}(o)-\theta c(o)\bigr],\\
\pi_{t,i}(\theta)
&=u_{t,i}(o^\star_{t,i})-\theta c(o^\star_{t,i}),
\end{aligned}
\label{eq:decision}
\end{equation}
where $u_{t,i}(o)$ is the estimated utility of option
$o$, $c(o)$ its storage cost, and $\pi_{t,i}$ the
resulting surplus. For previously stored tokens,
only prefixes no longer than their current codes
are considered. Keeping a frame also requires
$\beta$ bits of metadata for its retrieval embedding,
timestamp, and novelty. A frame is therefore retained
only when its tokens' combined surplus exceeds this cost:
\begin{equation}
\label{eq:keep}
\text{keep frame }t\quad\Longleftrightarrow\quad
\sum_{i=1}^{N}\pi_{t,i}(\theta)>\theta\beta.
\end{equation}

We initialize $\theta=0$. Whenever new data would
exceed the storage budget $B$, we use bisection to
increase the price until memory usage falls below
$\eta B$, where $\eta<1$ leaves room for incoming
frames. Increasing the price favors shorter codes
or removal, so stored representations never require
re-encoding. At a fixed price, these decisions
maximize the estimated utility minus the
price-weighted storage cost over the available
coding options (see supplement).

\paragraph{Utility.}
Since future questions are unknown, we estimate
each token's general usefulness rather than its
relevance to a specific query. We define its utility as:
\begin{equation}
\label{eq:utility}
u_{t,i}(\varnothing)=0,\qquad
u_{t,i}(r)=\alpha_{t,i}\,\nu_t\,\bigl(e(r)-e_0\bigr),
\end{equation}
where $\alpha_{t,i}$ measures token importance,
$\nu_t$ measures frame novelty, and $e(r)$ is the
prefix value from \cref{eq:er}.
The quality floor $e_0$ prevents retaining
representations that preserve less PCA variance than $e_0$,
controlling the trade-off between temporal coverage
and representation fidelity.

We estimate token importance $\alpha_{t,i}$ from
the attention received in the last full-attention
vision-encoder layer~\cite{yang2024visionzip,shang2024prumerge},
normalized to have mean one per frame.
Frame novelty is measured using recent frame
embeddings~\cite{song2024moviechat,he2024malmm}:
\begin{equation}
\label{eq:novelty}
\nu_t=\max\left\{\nu_{\min},
1-\max_{h\in H_{t-1}}g_t^\top h\right\},
\end{equation}
where $H_{t-1}$ contains recent frame embeddings,
including those not retained in memory.
For the first frame, $\nu_1=1$.

\paragraph{Cost.}
Each retained token costs $c(r)=r+b_{\mathrm{tok}}$
bits, including $r$ bits for its code and
$b_{\mathrm{tok}}$ bits for its spatial position,
importance, and code length, with $c(\varnothing)=0$.
Each retained frame additionally requires $\beta$
bits for its retrieval embedding, timestamp, and
novelty. The stored memory usage is:
\begin{equation}
\label{eq:bits}
\operatorname{bits}(\mathcal{M}_t)
=\sum_{t'\,\mathrm{kept}}\left[
\beta+\sum_{i:o_{t',i}\ne\varnothing}
(r_{t',i}+b_{\mathrm{tok}})
\right].
\end{equation}

\subsection{Query-Conditioned Reading}
\label{sec:read}

To retrieve relevant evidence from memory, we adopt a
lightweight CLIP-style approach inspired by
Q-Frame~\cite{zhang2025qframe}, adapted to stored
representations where frames and tokens may have been
discarded or compressed.

\paragraph{Frame and Token Selection.}
Given a question $q$, the CLIP-style text encoder gives $g_q=\phi_{\mathcal{T}}(q)$. The stored frames are then ranked by cosine similarity $g_t^\top g_q$.
Within each frame, we select tokens by decreasing
importance until their cumulative importance reaches
a fraction $\rho$ of the frame's retained importance:
\begin{equation}
\label{eq:coverage}
\sum_{i\in S_t}\alpha_{t,i}
\geq \rho\sum_{i\in K_t}\alpha_{t,i},
\end{equation}
where $K_t$ denotes the retained tokens and $S_t$ the
smallest top-ranked subset satisfying this threshold.
This adapts the number of selected tokens to their
importance distribution and the evidence remaining
in each frame.

We process frames in relevance order until the readout
budget $L$ is reached, counting both visual and
formatting tokens. Finally, we decode the selected tokens,
restore their spatial positions, and pass the frames
in chronological order to the frozen VLM reader.

\section{Experiments}
\label{sec:exps}

\definecolor{memryblue}{HTML}{206A98}
\definecolor{baselinegray}{HTML}{62666A}
\definecolor{fmtorange}{HTML}{D9822B}
\definecolor{fmtpurple}{HTML}{8C5A9E}
\definecolor{fmtgreen}{HTML}{4E8A3E}
\providecommand{\ours}{\textcolor{memryblue}{\textsc{Memryoshka}}}

\begin{table*}[t]
  \centering
  \caption{\textbf{Comparison at matched memory and readout}.
    Each budget block compares methods under the specified
    memory storage and readout budgets.
    No-cap references use unrestricted memory storage
    but the same readout budget, and are excluded from ranking.
    The +4 recent frames variant always includes the four most recent frames in the readout.
    \textbf{Bold} and \underline{underlining} indicate
    the best and second-best results within each budget,
    respectively.
    $^\dagger$Requires access to the full past video.}
  \label{tab:main}
  \footnotesize
  \setlength{\tabcolsep}{5pt}
  \renewcommand{\arraystretch}{1.08}
  \resizebox{\textwidth}{!}{%
  \begin{tabular}{l|cccccc|c}
    \toprule
    \textbf{Method}
      & \textbf{LVBench}
      & \textbf{LongVideoBench}
      & \textbf{Video-MME}
      & \textbf{TimeScope}
      & \textbf{OVO-Bench}
      & \textbf{StreamingBench}
      & \textbf{Avg.} \\
    \midrule

    \rowcolor{memsectiongray}
    \multicolumn{8}{c}{
      \textit{Memory Budget: 7.2 MB
      \quad | \quad Readout Budget: 1,000}
    } \\

    Base model (Full-Memory)
      & 44.5 & 54.4 & 49.6 & 85.6
      & 52.5 & 69.9 & 59.4 \\

    Uniform$^\dagger$
      & 30.1 & 41.8 & 43.6 & 27.8
      & 57.0 & 76.2 & 46.1 \\

      \hline

    SimpleStream~\cite{shen2026simplestream}
      & 27.2 & 39.9 & 41.5 & 24.4
      & 62.3 & 78.1 & 45.6 \\

    KeyRec~\cite{chen2026keyrec}
      & 27.2 & 41.0 & 39.8 & 24.4
      & \underline{62.4} & 77.5 & 45.4 \\

    CausalMem~\cite{song2026causalmem}
      & 33.4 & 46.6 & \underline{45.2}
      & 57.2 & 61.8 & 78.3 & 53.8 \\

    SAVEMem~\cite{wu2026savemem}
      & 27.9 & 36.2 & 40.0 & 26.1
      & \textbf{64.6}
      & \underline{79.8} & 45.8 \\

    \ours
      & \textbf{45.7}
      & \textbf{52.3}
      & \textbf{50.1}
      & \textbf{86.1}
      & 53.2 & 70.6
      & \underline{59.7} \\

    \rowcolor{memrowblue}
    \textbf{\ours{} + 4 recent frames}
      & \underline{45.5}
      & \underline{51.4}
      & \textbf{50.1}
      & \underline{84.4}
      & 60.3
      & \textbf{80.1}
      & \textbf{62.0} \\

    \midrule

    \rowcolor{memsectiongray}
    \multicolumn{8}{c}{
      \textit{Memory Budget: 14.3 MB
      \quad | \quad Readout Budget: 2,000}
    } \\

    Base model (Full-Memory)
      & 46.9 & 56.4 & 50.8 & 94.4
      & 54.9 & 72.9 & 62.7 \\

    Uniform$^\dagger$
      & 34.4 & 47.0 & 49.7 & 41.1
      & 57.4 & 77.3 & 51.2 \\

      \hline

    SimpleStream~\cite{shen2026simplestream}
      & 27.8 & 41.7 & 42.7 & 30.0
      & \underline{63.0} & 79.8 & 47.5 \\

    KeyRec~\cite{chen2026keyrec}
      & 27.9 & 42.9 & 41.3 & 24.4
      & 61.9 & 78.7 & 46.2 \\

    CausalMem~\cite{song2026causalmem}
      & 35.1 & 49.5 & 45.9 & 60.0
      & 62.7 & 80.3 & 55.6 \\

    SAVEMem~\cite{wu2026savemem}
      & 28.9 & 39.0 & 41.1 & 32.2
      & \textbf{64.2}
      & \underline{81.1} & 47.8 \\

    \ours
      & \textbf{48.4}
      & \textbf{58.7}
      & \underline{52.4}
      & \textbf{93.9}
      & 54.7 & 73.0
      & \underline{63.5} \\

    \rowcolor{memrowblue}
    \textbf{\ours{} + 4 recent frames}
      & \underline{47.2}
      & \underline{57.5}
      & \textbf{52.6}
      & \underline{88.9}
      & 59.7
      & \textbf{81.6}
      & \textbf{64.6} \\

    \midrule

    \rowcolor{memsectiongray}
    \multicolumn{8}{c}{
      \textit{Memory Budget: 28.7 MB
      \quad | \quad Readout Budget: 4,000}
    } \\

    Base model (Full-Memory)
      & 49.4 & 58.2 & 52.7 & 95.0
      & 56.6 & 76.0 & 64.6 \\

    Uniform$^\dagger$
      & 36.5 & 48.9 & 53.1 & 47.2
      & 59.4 & 79.4 & 54.1 \\

      \hline

    SimpleStream~\cite{shen2026simplestream}
      & 30.7 & 44.0 & 43.3 & 30.0
      & 61.2 & 80.2 & 48.2 \\

    KeyRec~\cite{chen2026keyrec}
      & 29.6 & 43.3 & 42.2 & 27.8
      & 61.8 & 79.9 & 47.4 \\

    CausalMem~\cite{song2026causalmem}
      & 35.6 & 52.3 & 45.4 & 64.4
      & \underline{62.4} & 80.7 & 56.8 \\

    SAVEMem~\cite{wu2026savemem}
      & 32.1 & 43.4 & 43.0 & 37.2
      & \textbf{64.5}
      & \underline{80.9} & 50.2 \\

    \ours
      & \underline{49.7}
      & \textbf{58.0}
      & \underline{53.7}
      & \underline{95.0}
      & 56.6 & 76.1
      & \underline{64.8} \\

    \rowcolor{memrowblue}
    \textbf{\ours{} + 4 recent frames}
      & \textbf{50.0}
      & \underline{56.9}
      & \textbf{54.2}
      & \textbf{95.6}
      & 59.4
      & \textbf{81.6}
      & \textbf{66.3} \\

    \bottomrule
  \end{tabular}%
  }
\end{table*}

\subsection{Experimental Setup}
\label{sec:setup}

\paragraph{Readers and Benchmarks.}
We use Qwen2.5-VL-7B~\cite{bai2025qwen25vl}
as the default frozen reader for experiments and ablations.
We evaluate on four long-video benchmarks:
LVBench~\cite{wang2024lvbench}, the long splits of
LongVideoBench~\cite{wu2024longvideobench},
the long-video subset of Video-MME~\cite{fu2024videomme}
(without subtitles), and TimeScope~\cite{zohar2025timescope}.
We also evaluate on two streaming benchmarks:
OVO-Bench~\cite{li2025ovobench} (backward-tracing
and real-time tasks) and StreamingBench~\cite{lin2024streamingbench}
(real-time visual understanding).
All memories are constructed without access to future questions.
Streaming questions access memory at their timestamps,
while offline questions are answered after processing
the full video.
The supplement provides benchmark details
and results across model sizes and architectures,
including LLaVA-OneVision~\cite{li2024llavaov}
and Gemma~\cite{gemma2026gemma4}.

\vspace{-0.7em}
\paragraph{Baselines.}
We compare against visual-memory methods:
SimpleStream~\cite{shen2026simplestream},
KeyRec~\cite{chen2026keyrec},
CausalMem~\cite{song2026causalmem}, and
SAVEMem~\cite{wu2026savemem}.
We also consider KV-cache-based methods:
InfiniPot-V~\cite{kim2025infinipotv},
StreamingTOM~\cite{chen2026streamingtom},
ReKV~\cite{di2025rekv},
StreamKV~\cite{chen2025streamkv}, and
MuKV~\cite{xiao2026mukv}.

\vspace{-0.7em}

\paragraph{Implementation Details.}
\label{sec:implementation}
By default, we sample frames at 1\,fps and resize them
to approximately $448\times448$ pixels while preserving
aspect ratio, yielding about 256 visual tokens per frame,
each of dimension $D=3{,}584$.
For retrieval, we use frozen LongCLIP-B/16~\cite{zhang2024longclip}
with square-padded frames.
We fit the PCA basis ($d=512$) on visual tokens from
50 randomly selected short and medium Video-MME videos.
Each PCA component is uniformly quantized to 8 bits
over $\pm3$ standard deviations.
We select the quality floor $e_0=0.5$ and readout share
$\rho=0.8$ using 600 questions from 200 short and medium
Video-MME videos.

The writer uses eight prefix lengths
$\mathcal{R}=\{32,64,128,\ldots,4{,}096\}$ bits,
with memory headroom $\eta=0.9$. Each token requires
$b_{\mathrm{tok}}=24$ bits of metadata for its
importance, spatial position, and code length.
Each retained frame requires $\beta=4{,}152$ bits
for its retrieval embedding, timestamp, and novelty.
Frame novelty uses the previous eight frames,
with $\nu_{\min}=0.05$.
Further implementation, evaluation, and efficiency
details are provided in the supplement.

\subsection{Main Results}
\label{sec:unlimited}

In our main experiments, we compare visual-memory methods
under matched storage and readout budgets.
Since some baselines pass their entire memory to the reader,
we set the storage budget $B$ to the nominal size of
$L$ full-precision visual tokens:
$L=\{1{,}000,2{,}000,4{,}000\}$ corresponds to
$B=\{7.2,14.3,28.7\}$\,MB.
We include two references: an uncapped memory that
stores all visual tokens but uses our query-conditioned
readout, and hindsight Uniform sampling, which accesses
the full video at query time.
Following SimpleStream~\cite{shen2026simplestream},
we additionally evaluate a Memryoshka variant with the four most
recent frames included in its readout.
\cref{tab:main} shows two main trends:

\begin{itemize}
\item \textbf{Long-video Understanding.}
Memryoshka outperforms all bounded visual-memory baselines
across four long-video benchmarks at every budget.
It also matches or outperforms the uncapped reference across the three budgets.
For perspective, one hour of video at 1\,fps requires
about 6.6\,GB of full-precision storage, compared with
just 7.2\,MB for Memryoshka.
Gains over CausalMem are particularly large on TimeScope
($28.9$--$33.9$ points) and LVBench ($12.3$--$14.1$ points),
where long-term evidence matters most.
Even at 7.2\,MB, Memryoshka matches or exceeds all
baselines at 28.7\,MB on these benchmarks.

\item \textbf{Streaming Understanding.}
On OVO-Bench and StreamingBench, methods favoring recent
frames perform better, suggesting that many questions
depend on recent observations as also observed by SimpleStream~\cite{shen2026simplestream}.
This also limits similarity-based retrieval in both
Memryoshka and the uncapped reference.
Simply adding the four most recent frames improves
Memryoshka by $2.8$--$9.5$ points across the two benchmarks,
achieving the best StreamingBench accuracy at every budget.
This variant also achieves the highest average across
all six benchmarks, exceeding CausalMem by
$8.2$--$9.5$ points.
\end{itemize}

\begin{figure}[t]
    \centering
    \begin{subfigure}[t]{\linewidth}
        \centering
        \includegraphics[width=\linewidth]{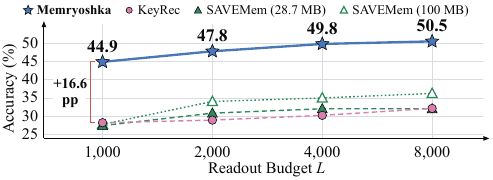}
        \caption{Decoupling Memory and Readout Budgets}
        \label{fig:readout_kv_a}
    \end{subfigure}

    \vspace{2mm}

    \begin{subfigure}[t]{\linewidth}
        \centering
        \includegraphics[width=\linewidth]{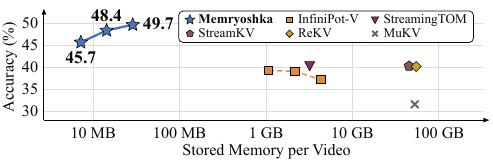}
        \caption{Comparison with KV-cache Memories}
        \label{fig:readout_kv_b}
    \end{subfigure}

    \caption{
\textbf{Memryoshka enables accurate long-video reasoning
with limited readout and compact storage.}
On Full LVBench with Qwen2.5-VL-7B,
(a) Memryoshka consistently outperforms KeyRec and SAVEMem
across readout budgets, demonstrating its effectiveness
under constrained readout.
(b) Compared with KV-cache baselines,
Memryoshka achieves higher accuracy while requiring
orders of magnitude less storage.
}

\vspace{-1em}
    \label{fig:readout_kv}
\end{figure}

\subsection{Decoupling Memory and Readout Budgets}
\label{sec:limited}

In \cref{sec:unlimited}, we match the storage budget to
the full-precision readout size.
Here, we instead fix the storage budget $B$ and vary
the readout budget $L$.
KeyRec~\cite{chen2026keyrec} and SAVEMem~\cite{wu2026savemem}
also support separate storage and readout budgets,
making them natural baselines for this comparison.
Unlike these methods, Memryoshka additionally adapts
the precision of stored evidence.

\cref{fig:readout_kv_a} compares these methods on LVBench
with a fixed 100\,MB storage budget and readout sizes
from 1{,}000 to 8{,}000 positions.
Memryoshka improves from 44.9\% to 50.5\%, while
KeyRec and SAVEMem remain below 33\%.
The released SAVEMem implementation limits storage
to approximately 28\,MB; we also remove this limit
to allow 100\,MB, but accuracy reaches only 36.3\%.
These results show that Memryoshka makes substantially
better use of the available memory and readout budgets.

\begin{table*}[t]
  \centering

\caption{\textbf{Temporal coverage vs. fidelity.}
LVBench results with a 1{,}000-position readout.
All use the same writer.
\emph{Frames} is the percentage of retained frames,
and \emph{Tok.} is average tokens per retained frame.
\emph{Oracle} uses annotated evidence timestamps,
while \emph{LongCLIP} uses query similarity; both report QA accuracy (\%).
For Memryoshka, bits/token gives the range of average code lengths
across budgets. \textbf{Bold} marks the best accuracy.
Memryoshka adapts precision across budgets,
matching or exceeding the best fixed code in five of six settings.}

  \label{tab:tradeoff}
  \footnotesize
  \setlength{\tabcolsep}{4pt}
  \renewcommand{\arraystretch}{1.08}
  \resizebox{\textwidth}{!}{%
  \begin{tabular}{lc|cccc|cccc|cccc}
    \toprule
    \rowcolor{memsectiongray}
    \textbf{Representation} & \textbf{Bits/token}
      & \multicolumn{4}{c|}{\textbf{Storage cap: 1.8\,MB}}
      & \multicolumn{4}{c|}{\textbf{Storage cap: 7.2\,MB}}
      & \multicolumn{4}{c}{\textbf{Storage cap: 28.7\,MB}} \\
    &
      & \multicolumn{2}{c}{\emph{Retained}}
      & \multicolumn{2}{c|}{\emph{Retrieval}}
      & \multicolumn{2}{c}{\emph{Retained}}
      & \multicolumn{2}{c|}{\emph{Retrieval}}
      & \multicolumn{2}{c}{\emph{Retained}}
      & \multicolumn{2}{c}{\emph{Retrieval}} \\
    \cmidrule(lr){3-4}\cmidrule(lr){5-6}
    \cmidrule(lr){7-8}\cmidrule(lr){9-10}
    \cmidrule(lr){11-12}\cmidrule(lr){13-14}
    & & Frames & Tok. & Oracle & LongCLIP
      & Frames & Tok. & Oracle & LongCLIP
      & Frames & Tok. & Oracle & LongCLIP \\
    \midrule
    \rowcolor{memsectiongray}
    \multicolumn{14}{c}{\textit{Fixed-rate PCA codes}} \\
    \quad 512 components $\times$ 8 bits (full code) & 4{,}096
      & 6.5 & 11 & 39.6 & 36.8
      & 18.8 & 16 & 46.7 & 41.5
      & 61.0 & 20 & 53.9 & 44.2 \\
    \quad 256 components $\times$ 8 bits & 2{,}048
      & 8.1 & 18 & 41.2 & 38.1
      & 24.7 & 24 & 48.8 & 42.0
      & 78.1 & 31 & 56.8 & 45.9 \\
    \quad 512 components $\times$ 2 bits & 1{,}024
      & 9.8 & 28 & 45.7 & 39.1
      & 33.2 & 34 & 51.9 & 44.1
      & 90.6 & 52 & 59.1 & 46.3 \\
    \quad 64 components $\times$ 8 bits & 512
      & 12.0 & 43 & 38.3 & 34.6
      & 45.0 & 48 & 42.1 & 35.7
      & 98.3 & 93 & 44.6 & 36.2 \\
    \midrule
    \rowcolor{memsectiongray}
    \multicolumn{14}{c}{\textit{Nested codes (fixed and adaptive prefix lengths)}} \\
    \quad Fixed $r=2{,}048$ & 2{,}048
      & 8.1 & 18 & 42.4 & 38.5
      & 24.7 & 24 & 50.6 & 41.8
      & 78.1 & 31 & 57.5 & 46.3 \\
    \quad Fixed $r=1{,}024$ & 1{,}024
      & 9.8 & 28 & 46.6 & 40.3
      & 33.2 & 34 & 53.8 & 44.0
      & 90.6 & 52 & \textbf{60.1} & 47.0 \\
    \quad Fixed $r=512$ & 512
      & 12.0 & 43 & 49.0 & 43.0
      & 45.0 & 48 & \textbf{55.8} & 44.1
      & 98.3 & 93 & 59.9 & 45.3 \\
    \quad Fixed $r=256$ & 256
      & 15.3 & 63 & 47.9 & 40.6
      & 58.4 & 66 & 54.6 & 42.7
      & 100.0 & 161 & 56.6 & 42.2 \\
    \rowcolor{memrowblue}
    \textbf{\ours{} (adaptive)} & 545--624
      & 11.2 & 45 & \textbf{49.4} & \textbf{43.2}
      & 39.2 & 50 & \textbf{55.8} & \textbf{45.7}
      & 95.2 & 80 & 59.9 & \textbf{47.1} \\
    \bottomrule
  \end{tabular}%
  }
\end{table*}

\begin{figure*}[t]
  \centering
  \begin{subfigure}{\columnwidth}
    \includegraphics[width=\linewidth]{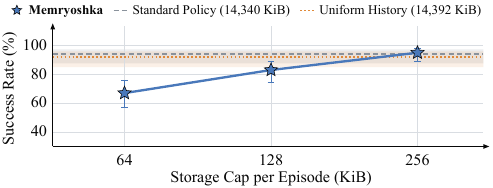}
    \caption{Robot policy: SimpleMemVLA on RMBench.}
    \label{fig:beyond_robot}
  \end{subfigure}\hfill
  \begin{subfigure}{\columnwidth}
    \includegraphics[width=\linewidth]{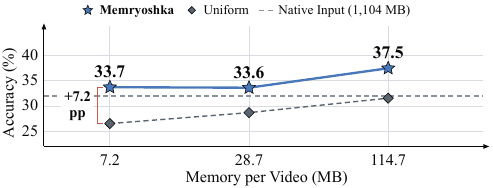}
    \caption{Audio and video: Qwen2.5-Omni-7B on LVOmniBench.}
    \label{fig:beyond_av}
  \end{subfigure}
  \caption{\textbf{Beyond video question answering.}
  (a) Success rate of SimpleMemVLA~\cite{yin2026simplememvla} with Memryoshka codes under per-episode storage caps (95\% confidence intervals).
  (b) Accuracy at matched nominal memory budgets; the native input uses its median memory footprint. Memryoshka maintains strong performance under high compression across both settings.}
  \label{fig:beyond}
  \vspace{-0.7em}
\end{figure*}

\subsection{Comparison with Streaming KV Memories}
\label{sec:kv}

We compare Memryoshka with streaming KV-cache
methods~\cite{di2025rekv,kim2025infinipotv,chen2026streamingtom,chen2025streamkv,xiao2026mukv}.
These methods process incoming frames through the LLM,
retaining KV states for reuse at question time.
For Qwen2.5-VL-7B, each full-precision KV-cache position
requires 57.3\,KB, compared with 7.2\,KB per visual token.
We evaluate all methods on LVBench using their
released implementations (\cref{fig:readout_kv_b}).
KV-cache methods achieve 31.6--40.4\% accuracy while
storing 1.1--54.6\,GB per video.
Memryoshka achieves 45.7--49.7\% with just
7.2--28.7\,MB, outperforming the strongest baseline
by 5.3 points with roughly 450$\times$ less storage.

Beyond memory efficiency, the methods differ in
when they incur computation.
KV-cache methods process every incoming frame through
the LLM, requiring substantial FLOPs and hundreds of
milliseconds per frame.
Memryoshka avoids LLM computation during memory writing
and instead prefills at most $L$ selected positions
when a question arrives.
While this incurs additional query-time prefill,
it substantially reduces streaming-time computation,
making Memryoshka better suited to long-running deployment.
Detailed compute and latency comparisons are provided
in the supplement.

\subsection{Coverage-Fidelity Trade-off}
\label{sec:tradeoff}

Under a fixed storage budget, greater temporal coverage
comes at the cost of representation fidelity.
We study this trade-off on LVBench,
whose annotated evidence timestamps allow us to
evaluate retrieval using the ground-truth evidence locations.
We compare fixed-rate PCA codes, fixed-length nested
codes, and Memryoshka's adaptive allocation, all using
the same writer. We report frame retention, tokens per
retained frame, and accuracy with both LongCLIP
and oracle retrieval, each under a 1{,}000-position
readout budget. The oracle uses evidence timestamps
to select among frames still available in memory,
helping separate retrieval errors from storage limitations.

\cref{tab:tradeoff} shows that the best fixed precision
depends on the storage budget. At 1.8\,MB, a full
4{,}096-bit PCA code retains only 6.5\% of frames
and achieves 36.8\% accuracy, while a 512-bit nested
code retains 12.0\% and reaches 43.0\%.
At 28.7\,MB, however, the best fixed length increases
to 1{,}024 bits.
Memryoshka achieves 43.2--47.1\% accuracy across
these budgets without choosing a fixed length,
matching or exceeding the best fixed code in five
of six budget--retrieval settings and trailing by
only 0.2 points in the sixth.
Similar results with oracle retrieval suggest that
the benefit comes from better allocation of stored
evidence, rather than retrieval quality alone.

\subsection{Beyond Video Question Answering}
\label{sec:robotics}

\paragraph{Robot policy.}
Some robot tasks require execution
history~\cite{chen2026rmbench}.
SimpleMemVLA~\cite{yin2026simplememvla} conditions on video
history and current observations, storing up to
23.4\,MiB of visual features for a 60-second window
and reading at most 4{,}800 visual tokens for prediction.
We integrate Memryoshka to compress memory and
retrieve evidence beyond this window.
We include a hindsight Uniform baseline
sampling 120 frames from the episode,
matching the model's visual context size.
\cref{fig:beyond_robot} compares them on 100
matched episodes across ten RMBench tasks~\cite{chen2026rmbench}.
With 256\,KiB of storage (56$\times$ less than the baseline),
Memryoshka achieves 95\% success, compared with 94\%
for SimpleMemVLA and 92\% for hindsight Uniform.
This shows that Memryoshka supports efficient robot
memories while maintaining strong performance under
high compression.

\vspace{-0.5em}

\paragraph{Audio-visual.}
We extend Memryoshka to audio-visual QA with
Qwen2.5-Omni-7B~\cite{xu2025qwen25omni} on
LVOmniBench~\cite{tao2026lvomnibench}.
We apply the same coding approach to audio, fitting
a separate PCA basis on short and medium Video-MME audio.
A shared price per bit allocates memory across both
modalities under a single budget.
For retrieval, we use LongCLIP vision-text embeddings
paired with the corresponding audio.
\cref{fig:beyond_av} compares Memryoshka with the
reported full-input baseline and Uniform sampling.
With just 7.2\,MB, Memryoshka achieves 33.7\% accuracy,
exceeding the full-input baseline's 32.0\% with
roughly 150$\times$ less storage (1.1\,GB per video).
At 114.7\,MB, the gain grows to 5.5 points.
Memryoshka also outperforms Uniform sampling at matched
storage by 4.9--7.2 points across budgets.

\subsection{Ablations}
\label{sec:ablations}

We examine the writer components and sensitivity to
the quality floor and readout coverage on LVBench
and the long split of Video-MME, at 1.8 and 7.2\,MB
with a 1{,}000-position readout (\cref{tab:ablation}).

\begin{table}[t]
  \centering
  \caption{\textbf{Ablations and parameter sensitivity.}
    QA accuracy (\%) on the long split of Video-MME and LVBench,
    with a 1{,}000-position readout.
    Each variant is compared with the default Memryoshka (blue).}
  \label{tab:ablation}
  \footnotesize
  \setlength{\tabcolsep}{4pt}
  \renewcommand{\arraystretch}{1.08}
  \resizebox{\columnwidth}{!}{%
  \begin{tabular}{l|cc|cc}
    \toprule
    \textbf{Configuration}
      & \multicolumn{2}{c|}{\textbf{Video-MME}}
      & \multicolumn{2}{c}{\textbf{LVBench}} \\
    \cmidrule(lr){2-3}\cmidrule(lr){4-5}
      & 1.8\,MB & 7.2\,MB & 1.8\,MB & 7.2\,MB \\
    \midrule
    \rowcolor{memrowblue}
    \textbf{\ours{} (default)} & 50.2 & 50.1 & 43.2 & 45.7 \\
    \midrule
    \rowcolor{memsectiongray}
    \multicolumn{5}{c}{\textit{Writer utility (component ablations)}} \\
    \quad Without importance ($\alpha_{t,i}=1$)
      & 48.1 & 49.3 & 39.1 & 43.4 \\
    \quad Without novelty ($\nu_t=1$)
      & 48.9 & 49.6 & 39.0 & 42.8 \\
    \quad Linear $e(r)$
      & 49.3 & 50.0 & 36.8 & 41.5 \\
    \midrule
    \rowcolor{memsectiongray}
    \multicolumn{5}{c}{\textit{Quality floor (sensitivity)}} \\
    \quad $e_0=0$   & 43.1 & 44.4 & 38.8 & 37.0 \\
    \quad $e_0=0.3$ & 48.9 & 48.8 & 41.0 & 45.6 \\
    \quad $e_0=0.7$ & 51.1 & 51.1 & 39.9 & 43.8 \\
    \quad $e_0=0.9$ & 50.7 & 50.0 & 38.9 & 41.5 \\
    \midrule
    \rowcolor{memsectiongray}
    \multicolumn{5}{c}{\textit{Readout coverage (sensitivity)}} \\
    \quad $\rho=0.5$ & 50.1 & 50.8 & 39.2 & 42.8 \\
    \quad $\rho=0.7$ & 50.1 & 49.8 & 42.4 & 43.4 \\
    \quad $\rho=1$   & 49.6 & 49.0 & 43.2 & 45.8 \\
    \bottomrule
  \end{tabular}%
  }
  \vspace{-1em}
\end{table}

\paragraph{Writer Utility.}
We ablate token importance $\alpha_{t,i}$, frame
novelty $\nu_t$, and the prefix value $e(r)$.
Removing importance or novelty reduces LVBench accuracy
by 2.3--4.2 points, showing the value of allocating
storage across both tokens and frames.
Replacing $e(r)$ with a linear function reduces
LVBench accuracy by 4.2--6.4 points, highlighting
the importance of valuing early bits more highly.
The effects are generally smaller on Video-MME.

\paragraph{Quality Floor.}
The floor $e_0$ balances coverage and fidelity.
Without it, more tokens are stored at very low
precision, but accuracy drops by 4.4--8.7 points.
Higher floors favor fidelity, slightly improving
Video-MME while reducing LVBench accuracy.
Our default $e_0=0.5$ is within 1.0 point of
the best tested value across all four settings.

\paragraph{Readout Coverage.}
The threshold $\rho$ controls how many tokens are
read per retrieved frame. Lower values allow more
frames but fewer tokens per frame: at $\rho=0.5$,
LVBench accuracy drops by 2.9--4.0 points.
Performance is otherwise fairly stable, with the
default $\rho=0.8$ within 1.0 point of the best
tested value in all four settings.

\section{Conclusion}
We studied the trade-off between temporal coverage and
representation fidelity in streaming VLM memory.
We introduced Memryoshka, a training-free approach
using nested token codes to adapt the precision of
stored visual evidence as the stream grows.
A shared price per bit jointly allocates memory across
frames, tokens, and precision, while a separate
readout retrieves relevant evidence when questions arrive.
Memryoshka approaches the performance of uncapped
full-precision memory at a fraction of the storage cost,
outperforming existing visual-memory baselines on long-video
understanding.
It also matches or outperforms the best fixed-precision
configurations in nearly all settings without choosing
a precision for each memory budget.
We further demonstrate its applicability to robot policies
and joint audio-visual memory.

Our results also highlight opportunities for improvement.
On streaming benchmarks, incorporating recent frames
improves performance, revealing a limitation of
similarity-based retrieval. Better retrieval could
improve the usefulness of stored evidence.
Similarly, our simple utility function provides an
effective proxy for memory allocation, but better
estimates of future usefulness could improve the
balance between temporal coverage and fidelity.

\section*{LLM Usage Statement}

ChatGPT and Claude were used to assist with manuscript preparation, including language editing, structural organization, refinement of mathematical notation, and the organization and presentation of tables and figures. These tools were also used for preliminary literature discovery to identify potentially relevant publications. The selection, examination, and interpretation of relevant literature were carried out by the authors. All references, technical claims, mathematical formulations, implementations, experimental results, and conclusions were independently reviewed and verified by the authors, who take full responsibility for the content of this submission.


{
    \small
    \bibliographystyle{ieeenat_fullname}
    \bibliography{main}

@String(CVPR= {IEEE Conf. Comput. Vis. Pattern Recog.})

@String(ICCV= {Int. Conf. Comput. Vis.})

@String(ECCV= {Eur. Conf. Comput. Vis.})

@String(NIPS= {Adv. Neural Inform. Process. Syst.})

@String(ICLR = {Int. Conf. Learn. Represent.})

@String(CVPR  = {CVPR})

@String(ICCV  = {ICCV})

@String(ECCV  = {ECCV})

@String(NIPS  = {NeurIPS})

@String(ICLR  = {ICLR})

@article{song2026causalmem,
  title   = {{Towards a Dynamic and Fixed-budget Memory Bank for Efficient Streaming Video Understanding}},
  author  = {Song, Baiyang and Lin, Yuli and Wu, Qiong and Chen, Tao and Peng, Jun and Chen, Xiao and others},
  journal = {arXiv preprint arXiv:2606.25658},
  year    = {2026}
}

@article{zhong2026prem,
  title   = {{PReM: Prefix-Steered Recurrent Memory for Long-Video Understanding}},
  author  = {Zhong, Siru and Wang, Qiongyan and Lv, Xiaohui and Zhuang, Yuzheng and Tao, Shuai and Liu, Wulong and others},
  journal = {arXiv preprint arXiv:2609.23601},
  year    = {2026}
}

@inproceedings{chen2026streamingtom,
  title     = {{StreamingTOM: Streaming Token Compression for Efficient Video Understanding}},
  author    = {Chen, Xueyi and Tao, Keda and Shao, Kele and Wang, Huan},
  booktitle = CVPR,
  year      = {2026}
}

@article{wang2025stc,
  title   = {{Accelerating Streaming Video Large Language Models via Hierarchical Token Compression}},
  author  = {Wang, Yiyu and Liu, Xuyang and Gui, Xiyan and Lin, Xinying and Yang, Boxue and Liao, Chenfei and others},
  journal = {arXiv preprint arXiv:2512.00891},
  year    = {2025}
}

@inproceedings{xie2026fluxmem,
  title     = {{FluxMem: Adaptive Hierarchical Memory for Streaming Video Understanding}},
  author    = {Xie, Yiweng and He, Bo and Wang, Junke and Zheng, Xiangyu and Ye, Ziyi and Wu, Zuxuan},
  booktitle = CVPR,
  year      = {2026}
}

@inproceedings{zeng2025streamforest,
  title     = {{StreamForest: Efficient Online Video Understanding with Persistent Event Memory}},
  author    = {Zeng, Xiangyu and Qiu, Kefan and Zhang, Qingyu and Li, Xinhao and Wang, Jing and Li, Jiaxin and others},
  booktitle = NIPS,
  year      = {2025}
}

@inproceedings{wang2025streambridge,
  title     = {{StreamBridge: Turning Your Offline Video Large Language Model into a Proactive Streaming Assistant}},
  author    = {Wang, Haibo and Feng, Bo and Lai, Zhengfeng and Xu, Mingze and Li, Shiyu and Ge, Weifeng and others},
  booktitle = NIPS,
  year      = {2025}
}

@inproceedings{xu2026streamingvlm,
  title     = {{StreamingVLM: Real-Time Understanding for Infinite Video Streams}},
  author    = {Xu, Ruyi and Xiao, Guangxuan and Chen, Yukang and He, Liuning and Lu, Yao and Han, Song},
  booktitle = ICLR,
  year      = {2026}
}

@inproceedings{wang2025videollamb,
  title     = {{VideoLLaMB: Long Streaming Video Understanding with Recurrent Memory Bridges}},
  author    = {Wang, Yuxuan and Song, Yiqi and Xie, Cihang and Liu, Yang and Zheng, Zilong},
  booktitle = ICCV,
  year      = {2025}
}

@inproceedings{li2026freshmem,
  title     = {{FreshMem: Brain-Inspired Frequency-Space Hybrid Memory for Streaming Video Understanding}},
  author    = {Li, Kangcong and Ye, Peng and Zhang, Lin and Wang, Chao and Qin, Huafeng and Fan, Jiayuan and others},
  booktitle = NIPS,
  year      = {2026}
}

@inproceedings{liu2024kivi,
  title     = {{KIVI: A Tuning-Free Asymmetric 2bit Quantization for KV Cache}},
  author    = {Liu, Zirui and Yuan, Jiayi and Jin, Hongye and Zhong, Shaochen and Xu, Zhaozhuo and Braverman, Vladimir and others},
  booktitle = {International Conference on Machine Learning (ICML)},
  year      = {2024}
}

@inproceedings{hooper2024kvquant,
  title     = {{KVQuant: Towards 10 Million Context Length LLM Inference with KV Cache Quantization}},
  author    = {Hooper, Coleman and Kim, Sehoon and Mohammadzadeh, Hiva and Mahoney, Michael W. and Shao, Yakun Sophia and Keutzer, Kurt and others},
  booktitle = NIPS,
  year      = {2024}
}

@article{lin2024matryoshkakv,
  title   = {{MatryoshkaKV: Adaptive KV Compression via Trainable Orthogonal Projection}},
  author  = {Lin, Bokai and Zeng, Zihao and Xiao, Zipeng and Kou, Siqi and Hou, Tianqi and Gao, Xiaofeng and others},
  journal = {arXiv preprint arXiv:2410.14731},
  year    = {2024}
}

@article{zhang2026rdkv,
  title   = {{RDKV: Rate-Distortion Bit Allocation for Joint Eviction and Quantization of the KV Cache}},
  author  = {Zhang, Junkai and Guo, Hang and Benini, Luca and Li, Yawei},
  journal = {arXiv preprint arXiv:2605.08317},
  year    = {2026}
}

@article{li2026breadthkv,
  title   = {{Spend Bytes on Breadth: Precision-Count Trade-offs for Decode-Time KV Compression in Long Chain-of-Thought Reasoning}},
  author  = {Li, Runguo},
  journal = {arXiv preprint arXiv:2610.05685},
  year    = {2026}
}

@article{dzikanyanga2026ttkv,
  title   = {{TTKV: Temporal-Tiered KV Cache for Long-Context LLM Inference}},
  author  = {Dzikanyanga, Gradwell and Yang, Weihao and Huang, Hao and Wu, Donglei and Wang, Shihao and Xia, Wen and others},
  journal = {arXiv preprint arXiv:2604.19769},
  year    = {2026}
}

@article{li2026sage,
  title   = {{SAGE: Efficient Protected Vision-Language Inference on Smart Glasses}},
  author  = {Li, Zhimin and Wang, Pan and Chen, Jingxian and Tang, Yuantao and Lou, Qian and Hu, Jingtong},
  journal = {arXiv preprint arXiv:2610.14089},
  year    = {2026}
}

@article{yuan2025tofc,
  title   = {{Task-Oriented Feature Compression for Multimodal Understanding via Device-Edge Co-Inference}},
  author  = {Yuan, Cheng and Liu, Zhening and Lv, Jiashu and Shao, Jiawei and Jiang, Yufei and Zhang, Jun and others},
  journal = {IEEE Transactions on Mobile Computing},
  year    = {2025}
}

@article{pang2026qtofc,
  title   = {{Task-Oriented Visual Feature Compression via Residual Vector Quantization for Device-Edge Multimodal Inference}},
  author  = {Pang, Luning and Yuan, Cheng and Shao, Jiawei and Huang, Mingtao and Shen, Yuan},
  journal = {arXiv preprint arXiv:2609.37090},
  year    = {2026}
}

@article{zhang2026rdtok,
  title   = {{Query Independent Variable Rate Visual Token Coding}},
  author  = {Zhang, Hongbo and Yang, Zihao and Song, Liuyang and Yang, Daqian and Yao, Haoyang and Wen, Yan and others},
  journal = {arXiv preprint arXiv:2610.00204},
  year    = {2026}
}

@inproceedings{kusupati2022matryoshka,
  title     = {{Matryoshka Representation Learning}},
  author    = {Kusupati, Aditya and Bhatt, Gantavya and Rege, Aniket and Wallingford, Matthew and Sinha, Aditya and Ramanujan, Vivek and others},
  booktitle = NIPS,
  year      = {2022}
}

@article{li2024llavaov,
  title   = {{LLaVA-OneVision}: Easy Visual Task Transfer},
  author  = {Li, Bo and Zhang, Yuanhan and Guo, Dong and Zhang, Renrui and Li, Feng and Zhang, Hao and others},
  journal = {arXiv preprint arXiv:2408.03326},
  year    = {2024}
}

@inproceedings{hayes2020remind,
  title     = {{REMIND Your Neural Network to Prevent Catastrophic Forgetting}},
  author    = {Hayes, Tyler L. and Kafle, Kushal and Shrestha, Robik and Acharya, Manoj and Kanan, Christopher},
  booktitle = ECCV,
  year      = {2020}
}

@inproceedings{wang2022mrdc,
  title     = {{Memory Replay with Data Compression for Continual Learning}},
  author    = {Wang, Liyuan and Zhang, Xingxing and Yang, Kuo and Yu, Longhui and Li, Chongxuan and Hong, Lanqing and others},
  booktitle = ICLR,
  year      = {2022}
}

@inproceedings{hu2026videommmu,
  title     = {{Video-MMMU: Evaluating Knowledge Acquisition from Multidisciplinary Professional Videos}},
  author    = {Hu, Kairui and Wu, Penghao and Pu, Fanyi and Xiao, Wang and Yue, Xiang and Li, Bo and Zhang, Yuanhan and Liu, Ziwei},
  booktitle = {ACL},
  pages     = {27798--27828},
  year      = {2026}
}

@misc{qwen2026qwen35,
  title        = {{Qwen3.5: Towards Native Multimodal Agents}},
  author       = {{Qwen Team}},
  howpublished = {\url{https://www.alibabacloud.com/blog/qwen3-5-towards-native-multimodal-agents_602894}},
  year         = {2026}
}

@inproceedings{lee2022dpict,
  title     = {{DPICT: Deep Progressive Image Compression Using Trit-Planes}},
  author    = {Lee, Jae-Han and Jeon, Seungmin and Choi, Kwang Pyo and Park, Youngo and Kim, Chang-Su},
  booktitle = CVPR,
  year      = {2022}
}

@article{li2023llamavid,
  title   = {{LLaMA-VID: An Image is Worth 2 Tokens in Large Language Models}},
  author  = {Li, Yanwei and Wang, Chengyao and Jia, Jiaya},
  journal = {arXiv preprint arXiv:2311.17043},
  year    = {2023}
}

@article{cai2024m3,
  title   = {{Matryoshka Multimodal Models}},
  author  = {Cai, Mu and Yang, Jianwei and Gao, Jianfeng and Lee, Yong Jae},
  journal = {arXiv preprint arXiv:2405.17430},
  year    = {2024}
}

@article{hu2024mqt,
  title   = {{Matryoshka Query Transformer for Large Vision-Language Models}},
  author  = {Hu, Wenbo and Dou, Zi-Yi and Li, Liunian Harold and Kamath, Amita and Peng, Nanyun and Chang, Kai-Wei},
  journal = {arXiv preprint arXiv:2405.19315},
  year    = {2024}
}

@article{yang2024mikv,
  title   = {{No Token Left Behind: Reliable KV Cache Compression via Importance-Aware Mixed Precision Quantization}},
  author  = {Yang, June Yong and Kim, Byeongwook and Bae, Jeongin and Kwon, Beomseok and Park, Gunho and Yang, Eunho and others},
  journal = {arXiv preprint arXiv:2402.18096},
  year    = {2024}
}

@article{tao2025vidkv,
  title   = {{Plug-and-Play 1.x-Bit KV Cache Quantization for Video Large Language Models}},
  author  = {Tao, Keda and You, Haoxuan and Sui, Yang and Qin, Can and Wang, Huan},
  journal = {arXiv preprint arXiv:2503.16257},
  year    = {2025}
}

@article{everett1963lagrange,
  title   = {Generalized {L}agrange Multiplier Method for Solving Problems of Optimum Allocation of Resources},
  author  = {Everett, Hugh},
  journal = {Operations Research},
  volume  = {11},
  number  = {3},
  pages   = {399--417},
  year    = {1963}
}

@article{bai2025qwen25vl,
  title   = {{Qwen2.5-VL Technical Report}},
  author  = {Shuai Bai and Keqin Chen and Xuejing Liu and Jialin Wang and Wenbin Ge and Sibo Song and others},
  journal = {arXiv preprint arXiv:2502.13923},
  year    = {2025}
}

@article{gemma2026gemma4,
  title   = {{Gemma 4 Technical Report}},
  author  = {{Gemma Team}},
  journal = {arXiv preprint arXiv:2607.02770},
  year    = {2026}
}

@inproceedings{radford2021clip,
  title     = {{Learning Transferable Visual Models From Natural Language Supervision}},
  author    = {Alec Radford and Jong Wook Kim and Chris Hallacy and Aditya Ramesh and Gabriel Goh and Sandhini Agarwal and Girish Sastry and Amanda Askell and Pamela Mishkin and Jack Clark and Gretchen Krueger and Ilya Sutskever},
  booktitle = {ICML},
  year      = {2021}
}

@inproceedings{chen2024fastv,
  title     = {{An Image is Worth 1/2 Tokens After Layer 2: Plug-and-Play Inference Acceleration for Large Vision-Language Models}},
  author    = {Liang Chen and Haozhe Zhao and Tianyu Liu and Shuai Bai and Junyang Lin and Chang Zhou and Baobao Chang},
  booktitle = ECCV,
  year      = {2024}
}

@article{yang2024visionzip,
  title   = {{VisionZip: Longer is Better but Not Necessary in Vision Language Models}},
  author  = {Senqiao Yang and Yukang Chen and Zhuotao Tian and Chengyao Wang and Jingyao Li and Bei Yu and Jiaya Jia},
  journal = {arXiv preprint arXiv:2412.04467},
  year    = {2024}
}

@article{jeddi2025saint,
  title   = {{Similarity-Aware Token Pruning: Your VLM but Faster}},
  author  = {Jeddi, Ahmadreza and Baghbanzadeh, Negin and Dolatabadi, Elham and Taati, Babak},
  journal = {arXiv preprint arXiv:2503.11549},
  year    = {2025}
}

@inproceedings{bolya2023tome,
  title     = {{Token Merging: Your ViT But Faster}},
  author    = {Daniel Bolya and Cheng-Yang Fu and Xiaoliang Dai and Peizhao Zhang and Christoph Feichtenhofer and Judy Hoffman},
  booktitle = ICLR,
  year      = {2023}
}

@inproceedings{shang2024prumerge,
  title     = {{LLaVA-PruMerge: Adaptive Token Reduction for Efficient Large Multimodal Models}},
  author    = {Yuzhang Shang and Mu Cai and Bingxin Xu and Yong Jae Lee and Yan Yan},
  booktitle = ICCV,
  year      = {2025}
}

@article{tao2024dycoke,
  title   = {{DyCoke: Dynamic Compression of Tokens for Fast Video Large Language Models}},
  author  = {Keda Tao and Can Qin and Haoxuan You and Yang Sui and Huan Wang},
  journal = {arXiv preprint arXiv:2411.15024},
  year    = {2024}
}

@article{shen2024longvu,
  title   = {{LongVU: Spatiotemporal Adaptive Compression for Long Video-Language Understanding}},
  author  = {Xiaoqian Shen and Yunyang Xiong and Changsheng Zhao and Lemeng Wu and Jun Chen and Chenchen Zhu and others},
  journal = {arXiv preprint arXiv:2410.17434},
  year    = {2024}
}

@article{yao2025timechatonline,
  title   = {{TimeChat-Online: 80\% Visual Tokens are Naturally Redundant in Streaming Videos}},
  author  = {Linli Yao and Yicheng Li and Yuancheng Wei and Lei Li and Shuhuai Ren and Yuanxin Liu and others},
  journal = {arXiv preprint arXiv:2504.17343},
  year    = {2025}
}

@inproceedings{fu2024framefusion,
  title     = {{FrameFusion: Combining Similarity and Importance for Video Token Reduction on Large Vision Language Models}},
  author    = {Tianyu Fu and Tengxuan Liu and Qinghao Han and Guohao Dai and Shengen Yan and Huazhong Yang and Xuefei Ning and Yu Wang},
  booktitle = ICCV,
  year      = {2025}
}

@inproceedings{tang2025aks,
  title     = {{Adaptive Keyframe Sampling for Long Video Understanding}},
  author    = {Xi Tang and Jihao Qiu and Lingxi Xie and Yunjie Tian and Jianbin Jiao and Qixiang Ye},
  booktitle = CVPR,
  year      = {2025}
}

@inproceedings{liu2025bolt,
  title     = {{BOLT: Boost Large Vision-Language Model Without Training for Long-form Video Understanding}},
  author    = {Shuming Liu and Chen Zhao and Tianqi Xu and Bernard Ghanem},
  booktitle = CVPR,
  year      = {2025}
}

@inproceedings{zhang2025qframe,
  title     = {{Q-Frame: Query-aware Frame Selection and Multi-Resolution Adaptation for Video-LLMs}},
  author    = {Shaojie Zhang and Jiahui Yang and Jianqin Yin and Zhenbo Luo and Jian Luan},
  booktitle = ICCV,
  year      = {2025}
}

@article{sun2025mdp3,
  title   = {{MDP3: A Training-free Approach for List-wise Frame Selection in Video-LLMs}},
  author  = {Hui Sun and Shiyin Lu and Huanyu Wang and Qing-Guo Chen and Zhao Xu and Weihua Luo and Kaifu Zhang and Ming Li},
  journal = {arXiv preprint arXiv:2501.02885},
  year    = {2025}
}

@inproceedings{yu2024framevoyager,
  title     = {{Frame-Voyager: Learning to Query Frames for Video Large Language Models}},
  author    = {Sicheng Yu and Chengkai Jin and Huanyu Wang and Zhenghao Chen and Sheng Jin and Zhongrong Zuo and Xiaolei Xu and Zhenbang Sun and Bingni Zhang and Jiawei Wu and Hao Zhang and Qianru Sun},
  booktitle = ICLR,
  year      = {2025}
}

@article{wang2024videoagent,
  title   = {{VideoAgent: Long-form Video Understanding with Large Language Model as Agent}},
  author  = {Xiaohan Wang and Yuhui Zhang and Orr Zohar and Serena Yeung-Levy},
  journal = {arXiv preprint arXiv:2403.10517},
  year    = {2024}
}

@inproceedings{song2024moviechat,
  title     = {{MovieChat: From Dense Token to Sparse Memory for Long Video Understanding}},
  author    = {Enxin Song and Wenhao Chai and Guanhong Wang and Yucheng Zhang and Haoyang Zhou and Feiyang Wu and others},
  booktitle = CVPR,
  year      = {2024}
}

@inproceedings{jeddi2026avis,
  title     = {{AVIS: Adaptive Test-Time Scaling for Vision-Language Models}},
  author    = {Jeddi, Ahmadreza and Le, Minh Ngoc and Kazerouni, Amirhossein and Karaimer, Hakki Can and Nguyen, Hue and Mohomed, Iqbal and others},
  booktitle = {Advances in Neural Information Processing Systems (NeurIPS)},
  year      = {2026}
}

@inproceedings{he2024malmm,
  title     = {{MA-LMM: Memory-Augmented Large Multimodal Model for Long-Term Video Understanding}},
  author    = {Bo He and Hengduo Li and Young Kyun Jang and Menglin Jia and Xuefei Cao and Ashish Shah and Abhinav Shrivastava and Ser-Nam Lim},
  booktitle = CVPR,
  year      = {2024}
}

@article{zhang2024flashvstream,
  title   = {{Flash-VStream: Memory-Based Real-Time Understanding for Long Video Streams}},
  author  = {Haoji Zhang and Yiqin Wang and Yansong Tang and Yong Liu and Jiashi Feng and Jifeng Dai and Xiaojie Jin},
  journal = {arXiv preprint arXiv:2406.08085},
  year    = {2024}
}

@article{qian2024videostreaming,
  title   = {{Streaming Long Video Understanding with Large Language Models}},
  author  = {Rui Qian and Xiaoyi Dong and Pan Zhang and Yuhang Zang and Shuangrui Ding and Dahua Lin and Jiaqi Wang},
  journal = {arXiv preprint arXiv:2405.16009},
  year    = {2024}
}

@inproceedings{chen2024videollmonline,
  title     = {{VideoLLM-online: Online Video Large Language Model for Streaming Video}},
  author    = {Joya Chen and Zhaoyang Lv and Shiwei Wu and Kevin Qinghong Lin and Chenan Song and Difei Gao and Jia-Wei Liu and Ziteng Gao and Dongxing Mao and Mike Zheng Shou},
  booktitle = CVPR,
  year      = {2024}
}

@article{qian2025dispider,
  title   = {{Dispider: Enabling Video LLMs with Active Real-Time Interaction via Disentangled Perception, Decision, and Reaction}},
  author  = {Rui Qian and Shuangrui Ding and Xiaoyi Dong and Pan Zhang and Yuhang Zang and Yuhang Cao and Dahua Lin and Jiaqi Wang},
  journal = {arXiv preprint arXiv:2501.03218},
  year    = {2025}
}

@inproceedings{di2025rekv,
  title     = {{Streaming Video Question-Answering with In-context Video KV-Cache Retrieval}},
  author    = {Shangzhe Di and Zhelun Yu and Guanghao Zhang and Haoyuan Li and Tao Zhong and Hao Cheng and Bolin Li and Wanggui He and Fangxun Shu and Hao Jiang},
  booktitle = ICLR,
  year      = {2025}
}

@inproceedings{kim2025infinipotv,
  title     = {{InfiniPot-V: Memory-Constrained KV Cache Compression for Streaming Video Understanding}},
  author    = {Minsoo Kim and Kyuhong Shim and Jungwook Choi and Simyung Chang},
  booktitle = NIPS,
  year      = {2025}
}

@article{yang2025streammem,
  title   = {{StreamMem: Query-Agnostic KV Cache Memory for Streaming Video Understanding}},
  author  = {Yanlai Yang and Zhuokai Zhao and Satya Narayan Shukla and Aashu Singh and Shlok Kumar Mishra and Lizhu Zhang and Mengye Ren},
  journal = {arXiv preprint arXiv:2508.15717},
  year    = {2025}
}

@inproceedings{xiao2026mukv,
  title     = {{MuKV: Multi-Grained KV Cache Compression for Long Streaming Video Question-Answering}},
  author    = {Junbin Xiao and Jiajun Chen and Tianxiang Sun and Xun Yang and Angela Yao},
  booktitle = CVPR,
  year      = {2026}
}

@article{chen2025streamkv,
  title   = {{StreamKV: Streaming Video Question-Answering with Segment-based KV Cache Retrieval and Compression}},
  author  = {Yilong Chen and Xiang Bai and Zhibin Wang and Chengyu Bai and Yuhan Dai and Ming Lu and Shanghang Zhang},
  journal = {arXiv preprint arXiv:2511.07278},
  year    = {2025}
}

@article{chen2026keyrec,
  title   = {{KeyRec: Bounded Visual Memory for Streaming and Long-Video Understanding}},
  author  = {Zihan Chen and Xuejian Rong and Xiaojuan Wang and Boqing Gong and Adi Zicher and Yael Pritch and Nikhil Karnad},
  journal = {arXiv preprint arXiv:2609.32182},
  year    = {2026}
}

@article{shen2026simplestream,
  title   = {{A Simple Baseline for Streaming Video Understanding}},
  author  = {Yujiao Shen and Shulin Tian and Jingkang Yang and Ziwei Liu},
  journal = {arXiv preprint arXiv:2604.02317},
  year    = {2026}
}

@article{wang2024lvbench,
  title   = {{LVBench: An Extreme Long Video Understanding Benchmark}},
  author  = {Weihan Wang and Zehai He and Wenyi Hong and Yean Cheng and Xiaohan Zhang and Ji Qi and Xiaotao Gu and Shiyu Huang and Bin Xu and Yuxiao Dong and Ming Ding and Jie Tang},
  journal = {arXiv preprint arXiv:2406.08035},
  year    = {2024}
}

@article{wu2024longvideobench,
  title   = {{LongVideoBench: A Benchmark for Long-context Interleaved Video-Language Understanding}},
  author  = {Haoning Wu and Dongxu Li and Bei Chen and Junnan Li},
  journal = {arXiv preprint arXiv:2407.15754},
  year    = {2024}
}

@article{fu2024videomme,
  title   = {{Video-MME: The First-Ever Comprehensive Evaluation Benchmark of Multi-modal LLMs in Video Analysis}},
  author  = {Chaoyou Fu and Yuhan Dai and Yongdong Luo and Lei Li and Shuhuai Ren and Renrui Zhang and others},
  journal = {arXiv preprint arXiv:2405.21075},
  year    = {2024}
}

@article{zhou2024mlvu,
  title   = {{MLVU: Benchmarking Multi-task Long Video Understanding}},
  author  = {Junjie Zhou and Yan Shu and Bo Zhao and Boya Wu and Zhengyang Liang and Shitao Xiao and Minghao Qin and Xi Yang and Yongping Xiong and Bo Zhang and Tiejun Huang and Zheng Liu},
  journal = {arXiv preprint arXiv:2406.04264},
  year    = {2024}
}

@inproceedings{li2025ovobench,
  title     = {{OVO-Bench: How Far is Your Video-LLMs from Real-World Online Video Understanding?}},
  author    = {Yifei Li and Junbo Niu and Ziyang Miao and Chunjiang Ge and Yuanhang Zhou and Qihao He and others},
  booktitle = CVPR,
  year      = {2025}
}

@article{lin2024streamingbench,
  title   = {{StreamingBench: Assessing the Gap for MLLMs to Achieve Streaming Video Understanding}},
  author  = {Junming Lin and Zheng Fang and Chi Chen and Zihao Wan and Fuwen Luo and Peng Li and Yang Liu and Maosong Sun},
  journal = {arXiv preprint arXiv:2411.03628},
  year    = {2024}
}

@article{yan2025teleego,
  title   = {{TeleEgo: Benchmarking Egocentric AI Assistants in the Wild}},
  author  = {Jiaqi Yan and Ruilong Ren and Jingren Liu and Shuning Xu and Ling Wang and Yiheng Wang and others},
  journal = {arXiv preprint arXiv:2510.23981},
  year    = {2025}
}

@article{shoham1988efficient,
  title={Efficient bit allocation for an arbitrary set of quantizers (speech coding)},
  author={Shoham, Yair and Gersho, Allen},
  journal={IEEE Transactions on Acoustics, Speech, and Signal Processing},
  volume={36},
  number={9},
  pages={1445--1453},
  year={1988},
  publisher={IEEE}
}

@article{taubman2000high,
  title={High performance scalable image compression with EBCOT},
  author={Taubman, David},
  journal={IEEE Transactions on image processing},
  volume={9},
  number={7},
  pages={1158--1170},
  year={2000},
  publisher={IEEE}
}

@inproceedings{xiao2024streamingllm,
  title     = {{Efficient Streaming Language Models with Attention Sinks}},
  author    = {Xiao, Guangxuan and Tian, Yuandong and Chen, Beidi and Han, Song and Lewis, Mike},
  booktitle = ICLR,
  year      = {2024}
}

@inproceedings{zhang2023h2o,
  title     = {{H$_2$O: Heavy-Hitter Oracle for Efficient Generative Inference of Large Language Models}},
  author    = {Zhang, Zhenyu and Sheng, Ying and Zhou, Tianyi and Chen, Tianlong and Zheng, Lianmin and Cai, Ruisi and others},
  booktitle = NIPS,
  year      = {2023}
}

@inproceedings{li2024snapkv,
  title     = {{SnapKV: LLM Knows What You are Looking for Before Generation}},
  author    = {Li, Yuhong and Huang, Yingbing and Yang, Bowen and Venkitesh, Bharat and Locatelli, Acyr and Ye, Hanchen and others},
  booktitle = NIPS,
  year      = {2024}
}

@inproceedings{he2024zipcache,
  title     = {{ZipCache: Accurate and Efficient KV Cache Quantization with Salient Token Identification}},
  author    = {He, Yefei and Zhang, Luoming and Wu, Weijia and Liu, Jing and Zhou, Hong and Zhuang, Bohan},
  booktitle = NIPS,
  year      = {2024}
}

@inproceedings{toderici2016variable,
  title     = {{Variable Rate Image Compression with Recurrent Neural Networks}},
  author    = {Toderici, George and O'Malley, Sean M. and Hwang, Sung Jin and Vincent, Damien and Minnen, David and Baluja, Shumeet and others},
  booktitle = ICLR,
  year      = {2016}
}

@article{nair2025matryoshkaquant,
  title   = {{Matryoshka Quantization}},
  author  = {Nair, Pranav and Datta, Puranjay and Dean, Jeff and Jain, Prateek and Kusupati, Aditya},
  journal = {arXiv preprint arXiv:2502.06786},
  year    = {2025}
}

@inproceedings{rippel2014nested,
  title     = {{Learning Ordered Representations with Nested Dropout}},
  author    = {Rippel, Oren and Gelbart, Michael A. and Adams, Ryan P.},
  booktitle = {International Conference on Machine Learning (ICML)},
  year      = {2014}
}

@inproceedings{wu2024videollmmod,
  title     = {{VideoLLM-MoD: Efficient Video-Language Streaming with Mixture-of-Depths Vision Computation}},
  author    = {Wu, Shiwei and Chen, Joya and Lin, Kevin Qinghong and Wang, Qimeng and Gao, Yan and Xu, Qianli and others},
  booktitle = NIPS,
  year      = {2024}
}

@article{shapiro1993ezw,
  title   = {{Embedded Image Coding Using Zerotrees of Wavelet Coefficients}},
  author  = {Shapiro, Jerome M.},
  journal = {IEEE Transactions on Signal Processing},
  volume  = {41},
  number  = {12},
  pages   = {3445--3462},
  year    = {1993}
}

@article{said1996spiht,
  title   = {{A New, Fast, and Efficient Image Codec Based on Set Partitioning in Hierarchical Trees}},
  author  = {Said, Amir and Pearlman, William A.},
  journal = {IEEE Transactions on Circuits and Systems for Video Technology},
  volume  = {6},
  number  = {3},
  pages   = {243--250},
  year    = {1996}
}

@article{lu2021plonq,
  title   = {{Progressive Neural Image Compression with Nested Quantization and Latent Ordering}},
  author  = {Lu, Yadong and Zhu, Yinhao and Yang, Yang and Said, Amir and Cohen, Taco S},
  journal = {arXiv preprint arXiv:2102.02913},
  year    = {2021}
}

@article{xiao2024infllm,
  title   = {{InfLLM: Training-Free Long-Context Extrapolation for LLMs with an Efficient Context Memory}},
  author  = {Xiao, Chaojun and Zhang, Pengle and Han, Xu and Xiao, Guangxuan and Lin, Yankai and Zhang, Zhengyan and others},
  journal = {arXiv preprint arXiv:2402.04617},
  year    = {2024}
}

@inproceedings{tang2024quest,
  title     = {{Quest: Query-Aware Sparsity for Efficient Long-Context LLM Inference}},
  author    = {Tang, Jiaming and Zhao, Yilong and Zhu, Kan and Xiao, Guangxuan and Kasikci, Baris and Han, Song},
  booktitle = {International Conference on Machine Learning (ICML)},
  year      = {2024}
}

@inproceedings{rae2020compressive,
  title     = {{Compressive Transformers for Long-Range Sequence Modelling}},
  author    = {Rae, Jack W. and Potapenko, Anna and Jayakumar, Siddhant M. and Lillicrap, Timothy P.},
  booktitle = ICLR,
  year      = {2020}
}

@article{wu2026savemem,
  title   = {{Semantic-Aware Adaptive Visual Memory for Streaming Video Understanding}},
  author  = {Hang Wu and Sherin Mary Mathews and Yujun Cai and Ming-Hsuan Yang and Yiwei Wang},
  journal = {arXiv preprint arXiv:2605.07897},
  year    = {2026}
}

@misc{zohar2025timescope,
  title        = {{TimeScope: How Long Can Your Video Large Multimodal Model Go?}},
  author       = {Orr Zohar and Rui Li and Andres Marafioti and Xiaohan Wang},
  howpublished = {\url{https://huggingface.co/blog/timescope-video-lmm-benchmark}},
  year         = {2025}
}

@article{xu2025qwen25omni,
  title   = {{Qwen2.5-Omni Technical Report}},
  author  = {Jin Xu and Zhifang Guo and Jinzheng He and Hangrui Hu and Ting He and Shuai Bai and Keqin Chen and Jialin Wang and Yang Fan and others},
  journal = {arXiv preprint arXiv:2503.20215},
  year    = {2025}
}

@article{zhang2026streamarena,
  title   = {{StreamArena: Toward Continuous, Interactive, and Long-Horizon Agentic Streaming Video Understanding}},
  author  = {Xichen Zhang and Guankai Li and Yinghao Zhu and Shijian Wang and Sitong Wu and Shaozuo Yu and Meng Chu and Yuan Lu and Jiaya Jia},
  journal = {arXiv preprint arXiv:2608.05703},
  year    = {2026}
}

@article{fox1966,
  author  = {Fox, Bennett},
  title   = {Discrete optimization via marginal analysis},
  journal = {Management Science},
  volume  = {13},
  number  = {3},
  pages   = {210--216},
  year    = {1966}
}

@inproceedings{zhang2024longclip,
  title     = {Long-{CLIP}: Unlocking the Long-Text Capability of {CLIP}},
  author    = {Zhang, Beichen and Zhang, Pan and Dong, Xiaoyi and Zang, Yuhang and Wang, Jiaqi},
  booktitle = {European Conference on Computer Vision (ECCV)},
  year      = {2024}
}

@article{chen2026rmbench,
  title   = {{RMBench: Memory-Dependent Robotic Manipulation Benchmark with Insights into Policy Design}},
  author  = {Chen, Tianxing and Wang, Yuran and Li, Mingleyang and Qin, Yan and Shi, Hao and Li, Zixuan and others},
  journal = {arXiv preprint arXiv:2603.01229},
  year    = {2026}
}

@article{yin2026simplememvla,
  title   = {{SimpleMemVLA: A Simple but Effective Native-Video Memory for Vision-Language-Action Models}},
  author  = {Yin, Cheng and Xu, Wang and Yang, Junpeng and Tam, Sikyuen and Liu, Hanyu and Yao, Yuan and others},
  journal = {arXiv preprint arXiv:2609.05533},
  year    = {2026}
}

@inproceedings{huang2025ovbench,
  title     = {{Online Video Understanding: OVBench and VideoChat-Online}},
  author    = {Huang, Zhenpeng and Li, Xinhao and Li, Jiaqi and Wang, Jing and Zeng, Xiangyu and Liang, Cheng and others},
  booktitle = CVPR,
  year      = {2025}
}

@article{tao2026lvomnibench,
  title={LVOmniBench: Pioneering Long Audio-Video Understanding Evaluation for Omnimodal LLMs},
  author={Tao, Keda and Zheng, Yuhua and Xu, Jia and Du, Wenjie and Shao, Kele and Wang, Hesong and Chen, Xueyi and Jin, Xin and Zhu, Junhan and Yu, Bohan and others},
  journal={arXiv preprint arXiv:2603.19217},
  year={2026}
}

@article{jeddi2026scopd,
  title   = {{SCOPD: Sparse-Context On-Policy Self-Distillation for Efficient Vision-Language Models}},
  author  = {Jeddi, Ahmadreza and Zhang, Enming and Gerigk, Jasper and Karaimer, Hakki Can and Nasr Azadani, Mozhgan and Luo, Jiayun and others},
  journal = {arXiv preprint arXiv:2609.34044},
  year    = {2026}
}
}

\clearpage
\appendix
\maketitlesupplementary
\section{Method Details}
\label{supp:method}

This section provides the derivations and implementation details of the three stages in Sec.~\ref{sec:method}: progressive token coding, online memory writing, and query-conditioned reading.

\subsection{Progressive Token Coding}
\label{supp:coding}

We detail the training-free PCA-based nested token code of Sec.~\ref{sec:codec}: how each component is quantized, how a prefix is decoded, why the bit ordering minimizes the expected reconstruction error at every prefix length, and what a prefix of each length contains.

\paragraph{Quantization.}
Each visual token $v\in\mathbb{R}^D$ is mapped to the standardized PCA coordinates $z=\bigl(W^\top(v-\mu)\bigr)\oslash\sigma$ of Eq.~\eqref{eq:codec_enc}, where $W\in\mathbb{R}^{D\times d}$ has orthonormal columns. For Qwen2.5-VL-7B, the $d=512$ components retain $90.1\%$ of the token variance around $\mu$. Each coordinate $z_k$ is uniformly quantized to an eight-bit level
\begin{equation}
j_k=\min\Bigl(255,\,\max\Bigl(0,\,\Bigl\lfloor \tfrac{256}{6}\,(z_k+3)\Bigr\rfloor\Bigr)\Bigr),
\label{eq:supp_level}
\end{equation}
written in binary from the most significant bit. The first $b$ bits of $j_k$ identify one of $2^b$ intervals of width $6/2^b$ on $[-3,3]$. The two outer intervals extend to $\pm\infty$, so values beyond three standard deviations remain covered. A full code contains $8d=4{,}096$ bits.

\paragraph{Decoding a Prefix.}
A prefix of $r$ bits retains the $b_k(r)$ most significant bits of each component $k$, where $b_k(r)$ follows from the fixed ordering. A stored code therefore needs only its length, which is part of the per-token metadata (Sec.~\ref{sec:write}), and no per-component allocation map. We reconstruct each component by the mean of a unit-variance Gaussian $\xi\sim\mathcal{N}(0,1)$ over its interval $[z_k^-,z_k^+)$:
\begin{equation}
\hat z_k=\mathbb{E}\bigl[\xi\mid z_k^-\le\xi<z_k^+\bigr]=\frac{n(z_k^-)-n(z_k^+)}{\Phi(z_k^+)-\Phi(z_k^-)},
\label{eq:supp_cond_mean}
\end{equation}
where $n$ and $\Phi$ denote the standard normal density and distribution function. Components without retained bits are set to $\hat z_k=0$, the PCA mean. The decoded token is
\begin{equation}
\hat v=\mu+W(\sigma\odot\hat z).
\label{eq:supp_decode}
\end{equation}
The interval means are computed once in closed form, so decoding requires one table lookup per component and one matrix product per token. It uses no learned parameters and applies to every prefix length. Shortening a stored code truncates its bit string, without access to the original frame or another pass of the vision encoder.

\paragraph{Value of a Bit Plane.}
Let $\psi_b(\xi)$ denote the reconstruction of $\xi\sim\mathcal{N}(0,1)$ from its first $b$ bits by Eq.~\eqref{eq:supp_cond_mean}, with $\psi_0(\xi)=0$. Since a conditional mean is orthogonal to its residual, the expected error after $b$ bits is
\begin{equation}
\varepsilon(b)=\mathbb{E}\bigl[(\xi-\psi_b(\xi))^2\bigr]=1-\mathbb{E}\bigl[\psi_b(\xi)^2\bigr],
\label{eq:supp_eps}
\end{equation}
and the expected reduction from bit $p$ is $\gamma(p)=\varepsilon(p-1)-\varepsilon(p)$. Tab.~\ref{tab:supp_gamma} lists the eight values. The first bit is the sign of $\xi$ and removes $\gamma(1)=2/\pi\approx0.64$ of the variance. The second and third bits remove $0.20$ and $0.12$, and each later bit removes roughly a quarter as much as the bit before it, as expected for a uniform quantizer at fine resolution. For component $k$, bit $p$ therefore reduces the expected squared error by $\sigma_k^2\gamma(p)$, the quantity that orders the bits in Sec.~\ref{sec:codec}.

\begin{table}[t]
\caption{\textbf{Value of each bit plane.} Expected reduction $\gamma(p)$ in squared error from bit $p$ and remaining error $\varepsilon(p)$ after $p$ bits, for a unit-variance Gaussian quantized to eight bits over $\pm3$ standard deviations. With interval means, the reductions decrease with $p$. With interval midpoints, the second bit is worth more than the first.}
\centering
\footnotesize
\begin{tabular}{@{}cccc@{}}
\toprule
 & \multicolumn{2}{c}{Interval mean} & Interval midpoint \\
\cmidrule(lr){2-3}\cmidrule(l){4-4}
Bit $p$ & $\gamma(p)$ & $\varepsilon(p)$ & $\gamma(p)$ \\
\midrule
1 & 0.6366 & 0.3634 & 0.1437 \\
2 & 0.2007 & 0.1627 & 0.6665 \\
3 & 0.1164 & 0.0463 & 0.1417 \\
4 & 0.0341 & 0.0122 & 0.0356 \\
5 & 0.0089 & 0.0033 & 0.0090 \\
6 & 0.0023 & 0.0010 & 0.0023 \\
7 & 0.0006 & 0.0004 & 0.0006 \\
8 & 0.0002 & 0.0003 & 0.0002 \\
\bottomrule
\end{tabular}

\label{tab:supp_gamma}
\end{table}

\paragraph{Optimality of the Ordering.}
Since $W$ has orthonormal columns, the error of a decoded token splits into an in-basis term and a residual that no code can recover:
\begin{equation}
\|\hat v-v\|^2=\sum_{k}\sigma_k^2(\hat z_k-z_k)^2+\|v_\perp\|^2,
\label{eq:supp_split}
\end{equation}
where $v_\perp=(I-WW^\top)(v-\mu)$ is the part of $v-\mu$ outside the basis.
Under the Gaussian PCA model of Sec.~\ref{sec:codec}, the expected in-basis error of a prefix that keeps $b_k$ bits of component $k$ is $\sum_k\sigma_k^2\varepsilon(b_k)$. Each term depends only on the distribution of its own component, since components are quantized and decoded separately. Sec.~\ref{sec:codec} states this result under an independent Gaussian model, but the decomposition needs only each component to be Gaussian, not the components to be independent.

\begin{proposition}[Prefix optimality]
\label{prop:supp_prefix}
List all pairs $(k,p)$ by decreasing $\sigma_k^2\gamma(p)$, breaking ties by smaller $k$ and then smaller $p$, and let $b_k(r)$ be the number of bits of component $k$ among the first $r$ pairs. If $\gamma(p)>0$ decreases with $p$, then for every $r\in\{0,\dots,8d\}$:
(i) the first $r$ pairs contain the $b_k(r)$ most significant bits of each component; and
(ii) $\bigl(b_k(r)\bigr)_k$ minimizes $\sum_k\sigma_k^2\varepsilon(b_k)$ over all allocations that keep the $b_k$ most significant bits of each component, with $\sum_k b_k=r$ and $0\le b_k\le8$.
\end{proposition}

\begin{proof}
(i) For a fixed $k$, $\sigma_k^2\gamma(p)$ decreases with $p$, so bit $p$ of component $k$ precedes bit $p+1$.
(ii) The objective is a sum of terms $\sigma_k^2\varepsilon(b_k)$, each decreasing and convex in $b_k$ because its successive decrements $\sigma_k^2\gamma(p)$ are positive and decreasing. For such separable objectives under a single budget, repeatedly adding the bit with the largest decrement gives an optimal allocation at every budget~\cite{fox1966,shoham1988efficient}. By (i), the next pair in the list is always the largest decrement still available, so the first $r$ pairs are exactly these greedy steps. Since each step extends the previous allocation, a single ordering serves all prefix lengths.
\end{proof}

Tab.~\ref{tab:supp_gamma} confirms that $\gamma(p)$ decreases for our quantizer, and our implementation checks this condition before building the ordering. The proposition concerns expected squared reconstruction error under the Gaussian model and the given quantizer; it does not guarantee optimal answer accuracy.

\paragraph{Role of Interval Means.}
The decreasing reductions in Tab.~\ref{tab:supp_gamma} rely on decoding with interval means. If a prefix were instead decoded with interval midpoints, the first bit would remove only $0.14$ of the variance and the second $0.67$. Sorting by $\sigma_k^2\gamma(p)$ would then place the second bit of a component before its first, and the greedy argument above would no longer apply. Interval means keep every prefix of the ordering a valid most-significant-bit prefix of each component.

\paragraph{Prefix Value and Contents.}
By Eq.~\eqref{eq:supp_eps}, the prefix value of Eq.~\eqref{eq:er} equals
\begin{equation}
e(r)=1-\frac{\sum_k\sigma_k^2\,\varepsilon\bigl(b_k(r)\bigr)}{\sum_k\sigma_k^2},
\label{eq:supp_er}
\end{equation}
the fraction of the in-basis variance that an $r$-bit prefix recovers under the Gaussian model. It is nondecreasing and concave in $r$, since the bits are sorted by decreasing reduction. Because the basis retains $90.1\%$ of the token variance, an $r$-bit prefix recovers about $0.901\,e(r)$ of the total. Tab.~\ref{tab:supp_prefix} lists $e(r)$ for the eight prefix lengths in $\mathcal{R}$ with the Qwen2.5-VL-7B basis, together with how each prefix spends its bits. Short prefixes spread their bits over many components at coarse precision rather than keeping a few components at full precision: a 512-bit prefix covers 326 of the 512 components and keeps 5 of the 8 bits of the leading component, and every component receives at least one bit from 1{,}024 bits on. This explains the gap in Tab.~\ref{tab:tradeoff} between a 512-bit nested code and 64 components at 8 bits, which use the same storage per token ($43.0\%$ vs.\ $34.6\%$ at 1.8\,MB with LongCLIP retrieval). The ordering and $e(r)$ are recomputed from the PCA basis of each reader.

\begin{table}[t]
\caption{\textbf{Contents of each prefix length.} Prefix value $e(r)$ (Eq.~\eqref{eq:supp_er}), number of components with at least one retained bit, and bits retained for the leading component, for the Qwen2.5-VL-7B PCA basis with $d=512$ components.}
\centering
\footnotesize
\begin{tabular}{@{}rccc@{}}
\toprule
$r$ (bits) & $e(r)$ & Components used & First-component bits \\
\midrule
32 & 0.317 & 25 & 3 \\
64 & 0.416 & 49 & 4 \\
128 & 0.533 & 94 & 4 \\
256 & 0.665 & 180 & 5 \\
512 & 0.804 & 326 & 5 \\
1{,}024 & 0.928 & 512 & 6 \\
2{,}048 & 0.993 & 512 & 7 \\
4{,}096 & 1.000 & 512 & 8 \\
\bottomrule
\end{tabular}
\label{tab:supp_prefix}
\end{table}

\subsection{Online Writing with a Shared Price}
\label{supp:write}

We detail the online writer of Sec.~\ref{sec:write}: what its decisions optimize at a fixed price, why raising the price never requires re-encoding, how the price is updated, and how utilities and storage are computed. Alg.~\ref{alg:supp_write} summarizes one write step.

\paragraph{Decisions at a Fixed Price.}
A memory $\mathcal{M}$ assigns each observed frame $t$ a decision to keep or discard it and, if the frame is kept, an option $o_{t,i}\in\{\varnothing\}\cup\mathcal{R}$ to each of its tokens. Its total utility is $U(\mathcal{M})=\sum_{t\,\mathrm{kept}}\sum_i u_{t,i}(o_{t,i})$, and its size $\operatorname{bits}(\mathcal{M})$ follows Eq.~\eqref{eq:bits}. Ties are resolved deterministically: a token is discarded when no option has a positive surplus, the shorter prefix wins between options of equal surplus, and a frame is discarded when its surplus equals its cost.

\begin{proposition}[Fixed-price optimality]
\label{prop:supp_price}
For any price $\theta\ge0$, keeping exactly the frames that satisfy Eq.~\eqref{eq:keep}, and storing each of their tokens with the option $o^\star_{t,i}(\theta)$ of Eq.~\eqref{eq:decision}, maximizes $U(\mathcal{M})-\theta\operatorname{bits}(\mathcal{M})$ over all memories formed from the observed frames. The resulting memory $\mathcal{M}^\star(\theta)$ has the highest utility among all memories of at most $\operatorname{bits}(\mathcal{M}^\star(\theta))$ bits.
\end{proposition}

\begin{proof}
The objective is a sum over frames. A discarded frame contributes zero. A kept frame contributes $\sum_i\bigl[u_{t,i}(o_{t,i})-\theta c(o_{t,i})\bigr]-\theta\beta$, in which each token term is maximized independently by $o^\star_{t,i}(\theta)$, giving $\sum_i\pi_{t,i}(\theta)-\theta\beta$. Keeping the frame is therefore optimal exactly when this value is positive, which is Eq.~\eqref{eq:keep}. The second claim follows from Everett's theorem~\cite{everett1963lagrange}: a memory with no more bits and a higher utility would also have a higher value of $U-\theta\operatorname{bits}$.
\end{proof}

Since $e(r)$ increases with $r$ and the cost $c(r)=r+b_{\mathrm{tok}}$ grows linearly, the chosen prefix length is a nondecreasing function of the ratio between the token weight $\alpha_{t,i}\nu_t$ and the price. At any price, important tokens in novel frames keep longer codes, and a higher price can only shorten them.

\paragraph{Raising the Price.}
The writer never lowers the price, and a price increase never requires the original frames:

\begin{proposition}[Price increases only truncate]
\label{prop:supp_monotone}
Let $\theta'\ge\theta$, and count $\varnothing$ as the shortest option.
(i) For every token, $o^\star_{t,i}(\theta')$ is no longer than $o^\star_{t,i}(\theta)$.
(ii) Every frame discarded at $\theta$ is also discarded at $\theta'$.
(iii) Re-deciding $\mathcal{M}^\star(\theta)$ at $\theta'$, with each stored token restricted to prefixes of its stored code, yields $\mathcal{M}^\star(\theta')$.
\end{proposition}

\begin{proof}
(i) Let $o=o^\star_{t,i}(\theta)$ and $o'=o^\star_{t,i}(\theta')$, and suppose $c(o')>c(o)$. Since the tie rule would choose the shorter $o$ at $\theta'$ on equal surplus, $o'$ must be strictly better there: $u(o')-u(o)>\theta'\bigl(c(o')-c(o)\bigr)\ge\theta\bigl(c(o')-c(o)\bigr)$. Then $o'$ is strictly better than $o$ at $\theta$ as well, which contradicts the choice of $o$.
(ii) Each surplus $\pi_{t,i}(\theta)$ is a maximum of functions that do not increase with $\theta$, so it does not increase, while $\theta\beta$ does not decrease. A frame with $\sum_i\pi_{t,i}(\theta)\le\theta\beta$ therefore also satisfies $\sum_i\pi_{t,i}(\theta')\le\theta'\beta$.
(iii) By (i), the option chosen at $\theta'$ is a prefix of the code stored at $\theta$, so restricting tokens to their stored prefixes changes no decision. By (ii), no discarded frame would be kept.
\end{proof}

Because the writer only raises the price, the memory at any time equals $\mathcal{M}^\star(\theta)$ at the current price, computed over all frames observed so far, even though discarded frames and truncated bits are never revisited. Every decision after a price increase is a truncation or a removal, which is why stored codes never need to be re-encoded.

\paragraph{Price Updates.}
The price starts at $\theta=0$, so early frames keep their full $4{,}096$-bit codes. Each incoming frame is decided at the current price, and stored frames are left unchanged. If adding the frame makes the serialized memory exceed $B$, the writer raises the price to the smallest value at which the memory fits within $\eta B$, with $\eta=0.9$. By Proposition~\ref{prop:supp_monotone}, the memory size does not increase with the price, so this value can be found by search. We start the upper end of the bracket at twice the current price, or at $10^{-12}$ while the price is zero, and double it until the memory fits. We then run 60 bisection steps on $\log\theta$ between the current price and this upper end, and keep the feasible end of the bracket. All stored frames, including the incoming one, are then re-decided at the new price, which only truncates or removes codes. The $10\%$ headroom lets the following frames be added at the same price until it is used up.

\paragraph{Utility Terms.}
The utility of Eq.~\eqref{eq:utility} combines three quantities.
\emph{Importance.} The vision encoder of Qwen2.5-VL uses windowed attention except in four full-attention layers. We recompute the attention probabilities of the last of these layers, sum the attention each patch receives over all query patches, average over heads, and average over the $2\times2$ patches merged into each visual token. The importance $\alpha_{t,i}$ is this value divided by its mean over the frame. It is stored as an 8-bit code of $\log_2\alpha_{t,i}$ over $[-8,8]$, and both the writer and the reader use the stored value.
\emph{Novelty.} Eq.~\eqref{eq:novelty} compares $g_t$ with the stored embeddings of the previous eight frames, whether or not they were kept. These eight embeddings form a small buffer that counts toward $B$, and $\nu_t$ is stored with 8 bits.
\emph{Quality floor.} A prefix can be retained only when $e(r)>e_0$. With $e_0=0.5$ and the Qwen2.5-VL-7B basis, this excludes the 32- and 64-bit prefixes (Tab.~\ref{tab:supp_prefix}), so no stored code is shorter than 128 bits.

\paragraph{Storage Accounting.}
Each retained token stores a 24-bit header, holding its 12-bit position in the merged grid, its 8-bit importance, and a 4-bit index of its prefix length, followed by its $r$-bit code, so $c(r)=r+24$. Each retained frame stores its retrieval embedding at 8 bits per dimension with a 16-bit scale, a 32-bit timestamp, and its 8-bit novelty, so $\beta=512\cdot8+16+32+8=4{,}152$ bits. Prices use this accounting, given by Eq.~\eqref{eq:bits}. The budget $B$, however, is enforced on the complete serialized memory, which also holds an 84-byte header, the novelty buffer of eight embeddings ($4{,}112$ bytes), and a 29-byte record header per kept frame with its index, grid size, token count, and record length. The reported budgets therefore bound all stored bytes. Propositions~\ref{prop:supp_price} and~\ref{prop:supp_monotone} refer to the size of Eq.~\eqref{eq:bits}, which leaves out these headers and the novelty buffer.

\begin{algorithm}[t]
\caption{Online memory writing for one frame.}
\label{alg:supp_write}
\begin{algorithmic}[1]
\Require memory $\mathcal{M}$ at price $\theta$; tokens $V_t$, embedding $g_t$, time $\tau_t$; budget $B$; headroom $\eta$
\State $\alpha_{t,i}\gets$ normalized last full-attention scores
\State $\nu_t\gets\max\{\nu_{\min},1-\max_{h\in H_{t-1}}g_t^\top h\}$; add $g_t$ to form $H_t$
\State encode each $v_{t,i}$ into its full $4{,}096$-bit code
\State decide frame $t$ at price $\theta$; add it to $\mathcal{M}$ if kept
\If{$\operatorname{size}(\mathcal{M})>B$}
\State $\theta\gets$ smallest price at which $\mathcal{M}$ fits $\eta B$
\State re-decide all stored frames at $\theta$
\EndIf
\State \Return $\mathcal{M},\theta$
\end{algorithmic}
\end{algorithm}

\subsection{Query-Conditioned Reading}
\label{supp:read}

We detail the readout of Sec.~\ref{sec:read}, summarized in Alg.~\ref{alg:supp_read}. Reading never modifies the memory, so later questions see the same stored evidence.

\paragraph{Frame Ranking.}
The LongCLIP text encoder embeds the question together with its answer options, written as the question followed by one line per option, with line breaks replaced by spaces and the text truncated to LongCLIP's 248-token context. Frame embeddings $g_t$ are computed from frames padded to a square with CLIP's mean color, so no part of a frame is cropped. Stored frames are ranked by the cosine similarity $g_t^\top g_q$ between their stored embeddings and the query embedding. The question $q$ in $g_q=\phi_{\mathcal{T}}(q)$ (Sec.~\ref{sec:read}) therefore always includes the answer options.

\paragraph{Token Selection and Budget.}
Frames are visited in rank order. Within a frame, stored tokens are sorted by decreasing stored importance, with ties broken by grid position, and the selected set $S_t$ is the shortest prefix of this order that satisfies Eq.~\eqref{eq:coverage}. Since $\alpha_{t,i}$ is normalized attention, $\rho$ is the share of the frame's retained attention that is read. Each frame also costs $f_t$ formatting positions for its timestamp text and two vision delimiters. For Qwen2.5-VL-7B, a timestamp such as ``[12:34]'' takes seven tokens, so $f_t=9$, or $11$ after the first hour. With $\ell$ positions already used, a frame contributes $\min(|S_t|,L-\ell-f_t)$ tokens, and reading stops once a further frame could not contribute a single token. The last frame read may therefore be cut to fit $L$. The question, its options, and the fixed prompt text are not counted in $L$.

\paragraph{Context Construction.}
The selected codes are decoded with Eq.~\eqref{eq:supp_decode}, and the frames are placed in chronological order. After the reader's prompt header and the line ``These are frames of a video.'', each frame contributes its timestamp text, a vision-start token, its decoded tokens in grid order, and a vision-end token. The question, its options, and the answer instruction follow. Each token keeps the rotary position of its original row and column in the frame's merged grid~\cite{bai2025qwen25vl}, and each frame advances the position index by the larger side of its full grid, as a complete frame would. Sparse frames therefore occupy the same positional extent as full frames, and the reader sees where each retained token came from.

\paragraph{Recent Frames.}
The +4 recent frames variant of Tab.~\ref{tab:main} places the four newest stored frames at the top of the ranking, ahead of the frames ranked by similarity. Token selection, budget accounting, and chronological presentation are unchanged, so the recent frames share the same readout budget $L$.

\begin{algorithm}[t]
\caption{Query-conditioned reading.}
\label{alg:supp_read}
\begin{algorithmic}[1]
\Require memory $\mathcal{M}_{t_q}$; question $q$ with options; budget $L$; share $\rho$
\State $g_q\gets\phi_{\mathcal{T}}(q\ \text{with options})$
\State rank stored frames by $g_t^\top g_q$
\State $\mathcal{C}\gets\emptyset$; $\ell\gets0$
\For{each stored frame $t$ in rank order}
\IfThen{$L-\ell-f_t<1$}{\textbf{break}}
\State $S_t\gets$ top tokens of frame $t$ covering share $\rho$
\State $S_t\gets$ first $\min(|S_t|,L-\ell-f_t)$ tokens of $S_t$
\State add frame $t$ with $S_t$ to $\mathcal{C}$; $\ell\gets\ell+f_t+|S_t|$
\EndFor
\State decode the codes in $\mathcal{C}$ and sort its frames by time
\State \Return $\mathcal{C}$ with timestamps and original grid positions
\end{algorithmic}
\end{algorithm}

\section{Experimental Setup}
\label{supp:setup}

This section details the implementation, the evaluation protocol, and the choice of hyperparameters for the experiments in Sec.~\ref{sec:exps}.

\subsection{Implementation Details}
\label{supp:impl}

\paragraph{Reader and Frames.}
We use the frozen Qwen2.5-VL-7B-Instruct reader~\cite{bai2025qwen25vl} in bfloat16 with scaled dot-product attention. Videos are decoded with PyAV and sampled causally at 1\,fps, so each sampled frame is the latest frame at or before its time and no future frame is used. Each frame is resized to an area of $448\times448$ pixels with its aspect ratio preserved and encoded on its own as an image, without merging consecutive frames into one temporal patch. A square frame gives $16\times16=256$ visual tokens and a 16:9 frame gives $12\times21=252$, each of dimension $D=3{,}584$, or $7{,}168$ bytes in bfloat16.

\paragraph{Token Codes.}
The PCA basis is fitted on visual tokens from 50 short and medium Video-MME videos~\cite{fu2024videomme}, using 40 evenly spaced frames from each. This gives $2{,}000$ frames and $504{,}000$ tokens, and the $d=512$ components retain $90.1\%$ of the token variance. Tokens are then coded as described in Sec.~\ref{supp:coding}.

\paragraph{Retrieval Embeddings.}
We use the official LongCLIP-B/16~\cite{zhang2024longclip} in fp32. Frames are padded to a square with CLIP's mean color and resized to $224\times224$, and the 512-dimensional unit-norm image embeddings are stored with 8 bits per dimension (Sec.~\ref{supp:write}). The text query contains the question and its options (Sec.~\ref{supp:read}).

\paragraph{Memory Settings.}
All settings follow Sec.~\ref{sec:implementation}: eight prefix lengths $\mathcal{R}=\{32,64,128,\ldots,4{,}096\}$, $e_0=0.5$, $\eta=0.9$, $b_{\mathrm{tok}}=24$, $\beta=4{,}152$, novelty over the previous eight frames with $\nu_{\min}=0.05$, and $\rho=0.8$. The storage budgets $B=7.2$, $14.3$, and $28.7$\,MB are exactly $7{,}168{,}000$, $14{,}336{,}000$, and $28{,}672{,}000$ bytes, the size of $1{,}000$, $2{,}000$, and $4{,}000$ full-precision tokens, with $1\,\text{MB}=10^6$ bytes. Likewise, the 1.8\,MB budget of Tabs.~\ref{tab:tradeoff} and~\ref{tab:ablation} and Sec.~\ref{supp:adaptivity} and the 0.9\,MB validation budget (Sec.~\ref{supp:hparams}) are $1{,}792{,}000$ and $896{,}000$ bytes ($250$ and $125$ tokens). The readout budget $L$ counts visual tokens and per-frame formatting positions (Sec.~\ref{supp:read}).

\paragraph{Frame Timestamps and Delimiters.}
For each kept frame, the memory stores its timestamp $\tau_t$ in seconds as a 32-bit float, counted in the per-frame cost $\beta$ (Sec.~\ref{supp:write}). At read time, the selected frames are placed in time order, and each timestamp is rounded to the nearest second and written as text, as \texttt{[mm:ss]} or, from the first hour on, as \texttt{[h:mm:ss]}. Each frame then enters the reader as its timestamp text, the vision-start token, its decoded visual tokens in grid order, and the vision-end token:
\begin{center}
\footnotesize\texttt{[00:12]<|vision\_start|>}$\,\cdots\,$\texttt{<|vision\_end|>[00:13]}$\,\cdots$
\end{center}
Consecutive frames follow each other without a separator, and a newline separates the last frame from the question. The Qwen2.5-VL tokenizer splits each digit into its own token, so a timestamp takes 7 tokens, or 9 from the first hour on, and each frame uses 9 or 11 formatting positions of the readout budget $L$ (Sec.~\ref{supp:read}).

\paragraph{Hardware and Software.}
Our runs use NVIDIA L40S GPUs (48\,GB) with PyTorch 2.13.0, Transformers 5.14.1, and PyAV 16.1.0.

\subsection{Evaluation Protocol}
\label{supp:protocol}

\paragraph{Question Timing.}
Memories are written without access to questions. For the streaming benchmarks, each question is answered from the memory written up to its timestamp, and questions asked at the same time share the same memory. For the other benchmarks, questions are answered after the full video has been written. Reading does not change the memory (Sec.~\ref{supp:read}).

\paragraph{Evidence Oracle.}
The oracle readout in Tab.~\ref{tab:tradeoff} uses the annotated evidence interval of each LVBench question, padded by one second on each side. Stored frames inside the interval are read first, in an order that spreads them over the interval. The remaining stored frames follow by their distance to the interval. Tokens are selected with the default rule ($\rho=0.8$) under the same budget $L=1{,}000$, and the frames are given to the reader in time order. Two of the $1{,}549$ questions have no annotated interval and use LongCLIP ranking instead.

\subsection{Hyperparameter Selection}
\label{supp:hparams}

We select the quality floor $e_0$ and readout share $\rho$ using a validation set of 600 questions from 200 short and medium Video-MME videos, with no video overlap with our evaluation sets. We select $e_0=0.5$ as it achieves the highest mean accuracy across memory budgets of 0.9 and 1.8\,MB among $\{0.3,0.4,0.5,0.6,0.7\}$, with all other values within one point. We select $\rho=0.8$ as it performs best among $\{0.7,0.8,0.9\}$.

\section{Baselines and Benchmarks}
\label{supp:benchmarks}

\subsection{Benchmarks}
\label{supp:bench}

Tab.~\ref{tab:supp_bench} summarizes the evaluation sets. We use the long split of
Video-MME~\cite{fu2024videomme} without subtitles, all LVBench~\cite{wang2024lvbench} tasks, the long split of the LongVideoBench~\cite{wu2024longvideobench}
validation set (videos of 15 to 60 minutes), the backward-tracing and real-time tasks
of OVO-Bench~\cite{li2025ovobench}, and the real-time visual understanding tasks of StreamingBench~\cite{lin2024streamingbench}.
For TimeScope~\cite{zohar2025timescope}, we take the first 15 questions of each task (temporal, QA, and OCR)
at four of its ten video lengths (1, 10, 60, and 180 minutes), one question per video.

\begin{table}[t]
\caption{\textbf{Evaluation sets.} Number of videos and questions, video length, and whether questions are asked after the full video or at their timestamps within the stream.}
\centering
\footnotesize
\setlength{\tabcolsep}{3pt}
\begin{tabular}{@{}lrrcc@{}}
\toprule
Benchmark & Videos & Questions & Minutes & Asked \\
\midrule
Video-MME & 300 & 900 & 30--60 & at end \\
LVBench & 103 & 1{,}549 & 30--140 & at end \\
LongVideoBench & 188 & 564 & 15--60 & at end \\
TimeScope & 180 & 180 & 1--180 & at end \\
OVO-Bench & 512 & 1{,}468 & 0.1--30 & in stream \\
StreamingBench & 500 & 2{,}500 & 1.5--23 & in stream \\
\bottomrule
\end{tabular}
\label{tab:supp_bench}
\end{table}

\begin{table*}[t]
  \centering
\caption{\textbf{Temporal coverage vs. fidelity with LLaVA-OneVision-7B.}
LVBench with $L=1{,}000$, following Tab.~\ref{tab:tradeoff}. Fixed rows retain $r$-bit nested codes. \emph{Frames} is the percentage of frames kept; \emph{Tok.} is the average tokens per kept frame. \emph{Oracle} ranks retained frames by annotated evidence timestamps, and \emph{LongCLIP} uses query similarity; both report QA accuracy (\%). For Memryoshka, bits/token gives the range of mean code lengths across budgets. \textbf{Bold} marks the best accuracy per column.}
  \label{tab:supp_llava_lvb}
  \footnotesize
  \setlength{\tabcolsep}{4pt}
  \renewcommand{\arraystretch}{1.08}
  \resizebox{\textwidth}{!}{%
  \begin{tabular}{lc|cccc|cccc|cccc}
    \toprule
    \rowcolor{memsectiongray}
    \textbf{Representation} & \textbf{Bits/token}
      & \multicolumn{4}{c|}{\textbf{Storage cap: 1.8\,MB}}
      & \multicolumn{4}{c|}{\textbf{Storage cap: 7.2\,MB}}
      & \multicolumn{4}{c}{\textbf{Storage cap: 28.7\,MB}} \\
    &
      & \multicolumn{2}{c}{\emph{Retained}}
      & \multicolumn{2}{c|}{\emph{Retrieval}}
      & \multicolumn{2}{c}{\emph{Retained}}
      & \multicolumn{2}{c|}{\emph{Retrieval}}
      & \multicolumn{2}{c}{\emph{Retained}}
      & \multicolumn{2}{c}{\emph{Retrieval}} \\
    \cmidrule(lr){3-4}\cmidrule(lr){5-6}
    \cmidrule(lr){7-8}\cmidrule(lr){9-10}
    \cmidrule(lr){11-12}\cmidrule(lr){13-14}
    & & Frames & Tok. & Oracle & LongCLIP
      & Frames & Tok. & Oracle & LongCLIP
      & Frames & Tok. & Oracle & LongCLIP \\
    \midrule
    \rowcolor{memsectiongray}
    \multicolumn{14}{c}{\textit{Nested codes (fixed and adaptive prefix lengths)}} \\
    \quad Fixed $r=4{,}096$ & 4{,}096
      & 11.0 & 6 & 30.6 & 31.2 & 41.9 & 7 & 33.8 & 33.9 & 99.3 & 12 & 41.8 & 35.2 \\
    \quad Fixed $r=2{,}048$ & 2{,}048
      & 16.1 & 8 & 32.8 & 32.3 & 61.9 & 8 & 35.8 & 32.9 & 100.0 & 24 & 48.8 & 41.1 \\
    \quad Fixed $r=1{,}024$ & 1{,}024
      & 22.5 & 10 & 34.5 & 32.4 & 78.9 & 12 & 39.5 & 35.1 & 100.0 & 47 & 53.7 & 42.2 \\
    \quad Fixed $r=512$ & 512
      & 29.5 & 13 & 36.1 & 33.1 & 90.0 & 20 & 43.6 & \textbf{40.1} & 100.0 & 91 & 57.6 & 42.8 \\
    \quad Fixed $r=256$ & 256
      & 36.1 & 18 & \textbf{39.0} & 34.3 & 95.4 & 35 & 46.4 & \textbf{40.1} & 100.0 & 146 & 56.6 & 43.3 \\
    \quad Fixed $r=128$ & 128
      & 40.0 & 26 & 36.7 & 34.4 & 98.1 & 61 & 46.8 & 38.1 & 100.0 & 190 & 52.9 & 40.3 \\
    \quad Fixed $r=64$ & 64
      & 41.6 & 42 & 35.3 & 32.9 & 99.1 & 98 & 40.5 & 35.2 & 100.0 & 196 & 42.6 & 35.3 \\
    \quad Fixed $r=32$ & 32
      & 42.2 & 64 & 30.2 & 30.1 & 99.3 & 130 & 31.4 & 30.0 & 100.0 & 196 & 32.3 & 29.6 \\
    \rowcolor{memrowblue}
    \ours{} (adaptive) & 171--266
      & 33.6 & 28 & 38.9 & \textbf{35.3} & 94.6 & 50 & \textbf{49.2} & 39.9 & 100.0 & 169 & \textbf{58.0} & \textbf{43.9} \\
    \bottomrule
  \end{tabular}}
\end{table*}

\begin{table*}[t]
  \centering
\caption{\textbf{Temporal coverage vs. fidelity on Video-MME.}
Long split without subtitles, $L=1{,}000$, and LongCLIP retrieval. \emph{Frames} and \emph{Tok.} describe stored memory as in Tab.~\ref{tab:tradeoff}; \emph{Acc.} is QA accuracy (\%). \textbf{Bold} marks the best accuracy per reader and budget.}
  \label{tab:supp_vmme}
  \footnotesize
  \setlength{\tabcolsep}{4pt}
  \renewcommand{\arraystretch}{1.08}
  \resizebox{\textwidth}{!}{%
  \begin{tabular}{lc|ccc|ccc|ccc}
    \toprule
    \rowcolor{memsectiongray}
    \textbf{Representation} & \textbf{Bits/token}
      & \multicolumn{3}{c|}{\textbf{Storage cap: 1.8\,MB}}
      & \multicolumn{3}{c|}{\textbf{Storage cap: 7.2\,MB}}
      & \multicolumn{3}{c}{\textbf{Storage cap: 28.7\,MB}} \\
    & & Frames & Tok. & Acc. & Frames & Tok. & Acc. & Frames & Tok. & Acc. \\
    \midrule
    \rowcolor{memsectiongray}
    \multicolumn{11}{c}{\textit{LLaVA-OneVision-7B}} \\
    \quad Fixed $r=4{,}096$ & 4{,}096
      & 17.4 & 7 & 42.2 & 71.8 & 6 & 40.9 & 100.0 & 20 & 44.7 \\
    \quad Fixed $r=2{,}048$ & 2{,}048
      & 25.9 & 8 & 40.9 & 94.9 & 9 & 42.6 & 100.0 & 40 & 44.8 \\
    \quad Fixed $r=1{,}024$ & 1{,}024
      & 37.1 & 10 & 40.0 & 99.9 & 17 & 45.1 & 100.0 & 79 & \textbf{45.4} \\
    \quad Fixed $r=512$ & 512
      & 48.4 & 13 & 40.9 & 100.0 & 33 & \textbf{45.2} & 100.0 & 154 & 43.2 \\
    \quad Fixed $r=256$ & 256
      & 59.0 & 18 & 41.6 & 100.0 & 63 & 45.1 & 100.0 & 196 & 44.8 \\
    \quad Fixed $r=128$ & 128
      & 66.1 & 26 & 41.0 & 100.0 & 116 & 42.6 & 100.0 & 196 & 42.0 \\
    \quad Fixed $r=64$ & 64
      & 69.9 & 39 & 39.7 & 100.0 & 178 & 39.7 & 100.0 & 196 & 40.2 \\
    \quad Fixed $r=32$ & 32
      & 72.1 & 58 & 37.9 & 100.0 & 196 & 36.1 & 100.0 & 196 & 36.1 \\
    \rowcolor{memrowblue}
    \ours{} (adaptive) & 169--403
      & 56.2 & 28 & \textbf{43.2} & 100.0 & 90 & 45.0 & 100.0 & 196 & 45.2 \\
    \midrule
    \rowcolor{memsectiongray}
    \multicolumn{11}{c}{\textit{Qwen2.5-VL-7B}} \\
    \quad Fixed $r=4{,}096$ & 4{,}096
      & 9.6 & 13 & 49.3 & 30.5 & 17 & 50.0 & 85.2 & 24 & \textbf{48.9} \\
    \quad Fixed $r=2{,}048$ & 2{,}048
      & 11.9 & 20 & \textbf{50.3} & 41.3 & 24 & \textbf{50.1} & 97.2 & 41 & 48.7 \\
    \quad Fixed $r=1{,}024$ & 1{,}024
      & 15.0 & 31 & \textbf{50.3} & 56.2 & 33 & \textbf{50.1} & 99.8 & 80 & 47.3 \\
    \quad Fixed $r=512$ & 512
      & 19.1 & 45 & 50.0 & 73.7 & 47 & 49.0 & 100.0 & 156 & 47.1 \\
    \quad Fixed $r=256$ & 256
      & 24.6 & 64 & 47.8 & 90.2 & 71 & 48.7 & 100.0 & 245 & 46.6 \\
    \quad Fixed $r=128$ & 128
      & 31.7 & 85 & 43.9 & 98.6 & 118 & 42.8 & 100.0 & 252 & 42.4 \\
    \rowcolor{memrowblue}
    \ours{} (adaptive) & 560--675
      & 17.3 & 47 & 50.2 & 65.2 & 49 & \textbf{50.1} & 100.0 & 120 & 48.3 \\
    \bottomrule
  \end{tabular}}
\end{table*}

\begin{table*}[t]
  \centering
\caption{\textbf{Temporal coverage vs. fidelity on StreamingBench.}
Real-time visual understanding, $L=1{,}000$. \emph{Frames} and \emph{Tok.} describe stored memory as in Tab.~\ref{tab:tradeoff}. \emph{LongCLIP} reports QA accuracy (\%); \emph{+4 recent} adds the four newest stored frames to the readout (Tab.~\ref{tab:main}). \textbf{Bold} marks the best per reader, budget, and readout.}
  \label{tab:supp_sb}
  \footnotesize
  \setlength{\tabcolsep}{2.5pt}
  \renewcommand{\arraystretch}{1.08}
  \resizebox{\textwidth}{!}{%
  \begin{tabular}{lc|cccc|cccc|cccc}
    \toprule
    \rowcolor{memsectiongray}
    \textbf{Representation} & \textbf{Bits/token}
      & \multicolumn{4}{c|}{\textbf{Storage cap: 1.8\,MB}}
      & \multicolumn{4}{c|}{\textbf{Storage cap: 7.2\,MB}}
      & \multicolumn{4}{c}{\textbf{Storage cap: 28.7\,MB}} \\
    & & Frames & Tok. & LongCLIP & +4 recent
      & Frames & Tok. & LongCLIP & +4 recent
      & Frames & Tok. & LongCLIP & +4 recent \\
    \midrule
    \rowcolor{memsectiongray}
    \multicolumn{14}{c}{\textit{LLaVA-OneVision-7B}} \\
    \quad Fixed $r=4{,}096$ & 4{,}096
      & 86.5 & 7 & 62.2 & 62.8 & 100.0 & 25 & 66.3 & 68.5 & 100.0 & 101 & 67.7 & 73.2 \\
    \quad Fixed $r=2{,}048$ & 2{,}048
      & 98.2 & 11 & 64.6 & 65.9 & 100.0 & 50 & 67.4 & 70.7 & 100.0 & 175 & \textbf{67.9} & 73.5 \\
    \quad Fixed $r=1{,}024$ & 1{,}024
      & 99.9 & 22 & 65.6 & 67.9 & 100.0 & 97 & \textbf{68.3} & 71.7 & 100.0 & 196 & 67.4 & 72.7 \\
    \quad Fixed $r=512$ & 512
      & 100.0 & 42 & 66.1 & 69.4 & 100.0 & 168 & 66.7 & 72.9 & 100.0 & 196 & 67.4 & 72.8 \\
    \quad Fixed $r=256$ & 256
      & 100.0 & 79 & 65.7 & 68.7 & 100.0 & 196 & 66.9 & 70.7 & 100.0 & 196 & 66.2 & 70.7 \\
    \quad Fixed $r=128$ & 128
      & 100.0 & 136 & 63.2 & 66.5 & 100.0 & 196 & 63.7 & 67.0 & 100.0 & 196 & 63.7 & 67.0 \\
    \quad Fixed $r=64$ & 64
      & 100.0 & 184 & 57.5 & 58.6 & 100.0 & 196 & 57.5 & 58.6 & 100.0 & 196 & 57.5 & 58.6 \\
    \quad Fixed $r=32$ & 32
      & 100.0 & 195 & 51.8 & 51.8 & 100.0 & 196 & 51.8 & 51.8 & 100.0 & 196 & 51.8 & 51.8 \\
    \rowcolor{memrowblue}
    \ours{} (adaptive) & 284--2{,}775
      & 100.0 & 113 & \textbf{66.9} & \textbf{71.0} & 100.0 & 196 & 68.2 & \textbf{73.3} & 100.0 & 196 & 67.7 & \textbf{73.9} \\
    \midrule
    \rowcolor{memsectiongray}
    \multicolumn{14}{c}{\textit{Qwen2.5-VL-7B}} \\
    \quad Fixed $r=4{,}096$ & 4{,}096
      & 54.6 & 11 & \textbf{71.7} & 73.0 & 97.6 & 26 & 71.0 & 76.4 & 100.0 & 102 & \textbf{70.3} & 79.9 \\
    \quad Fixed $r=2{,}048$ & 2{,}048
      & 67.9 & 17 & 71.5 & 74.9 & 99.7 & 49 & \textbf{71.2} & 78.4 & 100.0 & 190 & 70.2 & 80.1 \\
    \quad Fixed $r=1{,}024$ & 1{,}024
      & 81.1 & 27 & 71.0 & 75.1 & 100.0 & 98 & 70.3 & 79.1 & 100.0 & 249 & 69.5 & 79.2 \\
    \quad Fixed $r=512$ & 512
      & 90.9 & 47 & 71.3 & 75.9 & 100.0 & 180 & 70.6 & 78.7 & 100.0 & 252 & 69.6 & 78.8 \\
    \quad Fixed $r=256$ & 256
      & 97.2 & 83 & 69.0 & 75.3 & 100.0 & 246 & 68.1 & 76.7 & 100.0 & 252 & 68.1 & 76.7 \\
    \quad Fixed $r=128$ & 128
      & 99.6 & 142 & 63.2 & 68.6 & 100.0 & 252 & 62.7 & 69.3 & 100.0 & 252 & 62.7 & 69.3 \\
    \rowcolor{memrowblue}
    \ours{} (adaptive) & 709--2{,}330
      & 86.8 & 46 & \textbf{71.7} & \textbf{76.9} & 100.0 & 142 & 70.6 & \textbf{80.1} & 100.0 & 250 & 69.9 & \textbf{80.2} \\
    \bottomrule
  \end{tabular}}
\end{table*}
\begin{table*}[t]
  \centering
\caption{\textbf{Robustness across settings.}
Mean accuracy (\%) over 15 settings per reader: Video-MME; LVBench with LongCLIP or oracle retrieval; and StreamingBench with or without four recent frames, each at three storage caps. The reference is the best fixed length in each setting. \emph{Gap} and \emph{Worst} are the mean and maximum nonnegative shortfalls from that reference; \emph{Within 1} counts settings within one point.}
  \label{tab:supp_adapt_summary}
  \footnotesize
  \setlength{\tabcolsep}{4pt}
  \begin{tabular}{l|cc|cc|cccc}
    \toprule
    \rowcolor{memsectiongray}
    & \multicolumn{2}{c|}{\textbf{LLaVA-OV}} & \multicolumn{2}{c|}{\textbf{Qwen2.5-VL}} & \multicolumn{4}{c}{\textbf{All 30 settings}} \\
    \textbf{Memory} & Mean & Worst & Mean & Worst & Mean & Gap & Worst & Within 1 \\
    \midrule
    Fixed $r=4{,}096$ & 49.0 & 15.8 & 56.9 & 9.4 & 52.9 & 4.0 & 15.8 & 11/30 \\
    Fixed $r=2{,}048$ & 50.8 & 11.0 & 58.2 & 6.6 & 54.5 & 2.5 & 11.0 & 14/30 \\
    Fixed $r=1{,}024$ & 52.1 & 7.3 & 58.9 & 2.7 & 55.5 & 1.5 & 7.3 & 17/30 \\
    Fixed $r=512$ & 53.2 & 3.2 & 59.2 & 1.8 & 56.2 & 0.8 & 3.2 & 20/30 \\
    Fixed $r=256$ & 53.3 & 2.8 & 57.4 & 4.8 & 55.4 & 1.6 & 4.8 & 12/30 \\
    Fixed $r=128$ & 51.1 & 6.5 & 51.6 & 13.2 & 51.3 & 5.6 & 13.2 & 2/30 \\
    \rowcolor{memrowblue}
    \ours{} (adaptive) & \textbf{54.6} & \textbf{0.2} & \textbf{59.9} & \textbf{0.6} & \textbf{57.3} & \textbf{0.1} & \textbf{0.6} & \textbf{30/30} \\
    \bottomrule
  \end{tabular}
\end{table*}

\begin{table*}[t]
  \centering
\caption{\textbf{Storage and readout budgets on LVBench.}
LVBench QA accuracy (\%) across storage caps and readout budgets $L$ (positions). Results at $L=1{,}000$ repeat the corresponding code-length tables. \textbf{Bold} marks the best accuracy in each column.}
  \label{tab:supp_readout}
  \footnotesize
  \setlength{\tabcolsep}{4pt}
  \renewcommand{\arraystretch}{1.08}
  \resizebox{\textwidth}{!}{%
  \begin{tabular}{l|cccc|cccc|cccc}
    \toprule
    \rowcolor{memsectiongray}
    & \multicolumn{4}{c|}{\textbf{Storage cap: 1.8\,MB}} & \multicolumn{4}{c|}{\textbf{Storage cap: 7.2\,MB}} & \multicolumn{4}{c}{\textbf{Storage cap: 28.7\,MB}} \\
    \textbf{Memory} & 500 & 1{,}000 & 2{,}000 & 4{,}000 & 500 & 1{,}000 & 2{,}000 & 4{,}000 & 500 & 1{,}000 & 2{,}000 & 4{,}000 \\
    \midrule
    \rowcolor{memsectiongray}
    \multicolumn{13}{c}{\textit{Qwen2.5-VL-7B, LongCLIP retrieval}} \\
    \quad Fixed $r=4{,}096$ & 35.2 & 36.8 & 37.1 & 39.1 & 39.6 & 41.5 & 40.9 & 43.4 & 42.8 & 44.2 & 44.6 & 46.0 \\
    \quad Fixed $r=2{,}048$ & 36.7 & 38.5 & 40.1 & 40.9 & 41.8 & 41.8 & 44.5 & 44.2 & 43.4 & 46.3 & 46.5 & 48.0 \\
    \quad Fixed $r=1{,}024$ & 40.9 & 40.3 & 42.0 & 43.5 & 43.9 & 44.0 & 45.1 & 46.2 & 44.0 & 47.0 & 47.0 & 48.6 \\
    \quad Fixed $r=512$ & \textbf{41.7} & 43.0 & 43.4 & \textbf{44.8} & 43.7 & 44.1 & \textbf{45.5} & 46.8 & 42.8 & 45.3 & \textbf{47.5} & 48.9 \\
    \quad Fixed $r=256$ & 39.5 & 40.6 & 41.0 & 42.7 & 41.0 & 42.7 & 43.4 & 43.2 & 41.9 & 42.2 & 44.9 & 46.7 \\
    \quad Fixed $r=128$ & 37.4 & 37.7 & 37.3 & 37.9 & 36.9 & 38.7 & 39.3 & 39.0 & 37.1 & 37.0 & 38.4 & 40.6 \\
    \rowcolor{memrowblue}
    \quad \ours{} & 41.1 & \textbf{43.2} & \textbf{44.5} & 44.3 & \textbf{44.2} & \textbf{45.7} & 45.4 & \textbf{46.9} & \textbf{44.4} & \textbf{47.1} & 47.3 & \textbf{49.7} \\
    \midrule
    \rowcolor{memsectiongray}
    \multicolumn{13}{c}{\textit{Qwen2.5-VL-7B, oracle retrieval}} \\
    \quad Fixed $r=4{,}096$ & 37.9 & 39.6 & 39.1 & 38.8 & 46.4 & 46.7 & 46.3 & 44.0 & 52.4 & 53.9 & 54.4 & 54.6 \\
    \quad Fixed $r=2{,}048$ & 41.0 & 42.4 & 41.8 & 40.9 & 49.3 & 50.6 & 50.5 & 49.6 & 56.8 & 57.5 & 58.4 & 58.1 \\
    \quad Fixed $r=1{,}024$ & 45.2 & 46.6 & 45.2 & 43.8 & 52.7 & 53.8 & 53.6 & 53.0 & 57.6 & \textbf{60.1} & 60.7 & 59.5 \\
    \quad Fixed $r=512$ & \textbf{48.7} & 49.0 & \textbf{48.5} & \textbf{47.5} & \textbf{54.3} & \textbf{55.8} & \textbf{56.8} & 55.0 & 57.1 & 59.9 & \textbf{61.3} & 61.4 \\
    \quad Fixed $r=256$ & 47.7 & 47.9 & 46.4 & 45.7 & 52.4 & 54.6 & 54.0 & 52.2 & 53.3 & 56.6 & 58.4 & 58.1 \\
    \quad Fixed $r=128$ & 40.7 & 43.2 & 42.5 & 40.3 & 45.4 & 45.8 & 46.6 & 45.7 & 45.0 & 46.9 & 47.3 & 48.0 \\
    \rowcolor{memrowblue}
    \quad \ours{} & 48.4 & \textbf{49.4} & 47.8 & 46.9 & 53.9 & \textbf{55.8} & 56.4 & \textbf{55.1} & \textbf{58.5} & 59.9 & \textbf{61.3} & \textbf{61.9} \\
    \midrule
    \rowcolor{memsectiongray}
    \multicolumn{13}{c}{\textit{LLaVA-OneVision-7B, LongCLIP retrieval}} \\
    \quad Fixed $r=4{,}096$ & 31.3 & 31.2 & 31.5 & 31.5 & 32.0 & 33.9 & 33.3 & 32.9 & 35.9 & 35.2 & 36.5 & 35.7 \\
    \quad Fixed $r=2{,}048$ & 33.6 & 32.3 & 31.7 & 31.9 & 32.9 & 32.9 & 33.3 & 33.0 & 39.3 & 41.1 & 43.8 & 41.8 \\
    \quad Fixed $r=1{,}024$ & 33.3 & 32.4 & 33.4 & 32.7 & 34.9 & 35.1 & 35.3 & 35.6 & 41.3 & 42.2 & 44.4 & 43.7 \\
    \quad Fixed $r=512$ & 33.5 & 33.1 & 34.0 & 33.4 & 37.7 & \textbf{40.1} & 39.3 & 39.5 & 41.6 & 42.8 & 43.7 & 44.9 \\
    \quad Fixed $r=256$ & 34.2 & 34.3 & 33.7 & 34.5 & \textbf{39.1} & \textbf{40.1} & 40.9 & 41.1 & 42.3 & 43.3 & 44.4 & 45.7 \\
    \quad Fixed $r=128$ & 34.3 & 34.4 & 33.9 & 35.2 & 37.7 & 38.1 & 39.5 & 39.4 & 37.9 & 40.3 & 40.2 & 42.7 \\
    \rowcolor{memrowblue}
    \quad \ours{} & \textbf{34.8} & \textbf{35.3} & \textbf{35.5} & \textbf{35.3} & 38.6 & 39.9 & \textbf{42.4} & \textbf{41.2} & \textbf{42.5} & \textbf{43.9} & \textbf{44.5} & \textbf{46.4} \\
    \midrule
    \rowcolor{memsectiongray}
    \multicolumn{13}{c}{\textit{LLaVA-OneVision-7B, oracle retrieval}} \\
    \quad Fixed $r=4{,}096$ & 31.7 & 30.6 & 32.5 & 32.6 & 33.2 & 33.8 & 33.9 & 33.7 & 40.3 & 41.8 & 41.2 & 41.3 \\
    \quad Fixed $r=2{,}048$ & 34.0 & 32.8 & 32.5 & 33.3 & 34.5 & 35.8 & 35.7 & 35.5 & 47.5 & 48.8 & 48.6 & 46.9 \\
    \quad Fixed $r=1{,}024$ & 33.9 & 34.5 & 33.8 & 34.1 & 39.2 & 39.5 & 39.8 & 38.5 & 51.2 & 53.7 & 54.2 & 54.5 \\
    \quad Fixed $r=512$ & 35.1 & 36.1 & 35.3 & 35.7 & 43.4 & 43.6 & 45.4 & 44.2 & \textbf{55.7} & 57.6 & \textbf{57.8} & 55.6 \\
    \quad Fixed $r=256$ & 38.5 & \textbf{39.0} & 37.6 & 36.5 & 46.2 & 46.4 & \textbf{49.0} & 46.3 & 53.4 & 56.6 & 56.8 & 57.0 \\
    \quad Fixed $r=128$ & 36.6 & 36.7 & 37.6 & 37.7 & 45.2 & 46.8 & 47.8 & 45.0 & 49.9 & 52.9 & 51.4 & 50.6 \\
    \rowcolor{memrowblue}
    \quad \ours{} & \textbf{38.8} & 38.9 & \textbf{39.4} & \textbf{37.9} & \textbf{47.6} & \textbf{49.2} & 48.6 & \textbf{48.8} & 54.8 & \textbf{58.0} & \textbf{57.8} & \textbf{58.1} \\
    \bottomrule
  \end{tabular}}
\end{table*}

\subsection{Baselines}
\label{supp:baselines}

\paragraph{Shared Protocol.}
All methods use the same frozen reader, the same 1\,fps frames (Sec.~\ref{supp:impl}), and the same visual tokens, computed once per frame through Qwen's image path. Every method answers the same question text, including the answer options. For each baseline, we run the authors' released code, adapted to the shared reader, with the settings of their official scripts; all numbers come from our runs, not from the original papers. Each baseline keeps its own memory, selection rule, prompt, and answer extraction, and decodes greedily. On TimeScope, all methods use the official prompt and scorer.

\paragraph{References.}
Uniform, our hindsight reference, has access to the full past video at question time. It reads uniformly spaced full-precision frames that fit within $B$ and $L$ (up to 3, 7, and 15 frames at the three budgets) and answers exactly as Memryoshka does. The uncapped reference stores all visual tokens and reads them with Memryoshka's readout.

\paragraph{Visual-Memory Baselines.}
Each baseline's full persistent state counts against the storage budget $B$.
\begin{itemize}
\item SimpleStream~\cite{shen2026simplestream} keeps the newest whole frames that fit $B$ and reads them in time order.
\item CausalMem~\cite{song2026causalmem} keeps a question-independent token memory and an online basis, both counted toward $B$.
\item KeyRec~\cite{chen2026keyrec} keeps a cache of recent tokens and a fixed number of key events, sized to fit $B$. At question time, it divides its readout between events and recent tokens.
\item SAVEMem~\cite{wu2026savemem} keeps short-, mid-, and long-term memories and retrieves from them using the question.
\end{itemize}
Baselines use their own readout rules, which can read fewer than $L$ tokens; in Tab.~\ref{tab:main}, KeyRec's readout is always shorter than $L$. We change these readouts in only two ways: CausalMem's internal cap on readable tokens is raised so that only $L$ limits its readout, and a SAVEMem selection longer than $L$ is cut to fit, keeping its short-term frames and adding retrieved frames by relevance until $L$ is reached. For Fig.~\ref{fig:readout_kv_a}, we also remove SAVEMem's internal storage cap, so that its state is
bounded only by the 100\,MB budget. At 100\,MB, KeyRec's native readout is
6{,}412 tokens, longer than $L$ for $L \le 4{,}000$; it is then cut to $L$,
keeping whole event segments in KeyRec's order and filling the rest with the
newest recent tokens.

\paragraph{KV-Cache Baselines.}
We run ReKV~\cite{di2025rekv}, StreamingTOM~\cite{chen2026streamingtom}, StreamKV~\cite{chen2025streamkv}, MuKV~\cite{xiao2026mukv}, and InfiniPot-V~\cite{kim2025infinipotv} from their official releases on the same reader and frames, keeping their released settings except where Qwen's larger frames change a token count. ReKV keeps the full KV cache of every frame, and StreamingTOM, StreamKV, and MuKV compress it as frames arrive; none of these four bounds its memory. InfiniPot-V is run at its 1, 2, and 4\,GiB budgets ($1\,\text{GiB}=2^{30}$ bytes). Stored memory is every byte a method holds when the questions arrive, including retrieval keys, prompt prefix, and local window for the KV-cache methods, reported as the median over the 103 LVBench videos. Memryoshka uses 7.2, 14.3, and 28.7\,MB with $L=1{,}000$, $2{,}000$, and $4{,}000$.

\section{Adaptive versus Fixed Code Lengths}
\label{supp:adaptivity}

Memryoshka allocates memory across frames, tokens, and prefix lengths using a shared price per bit (Secs.~\ref{sec:write} and~\ref{supp:write}). We compare it with fixed-length writers under the same allocation rule across two readers, three benchmarks, and three storage budgets.

\subsection{Fixed-Length Writers}
\label{supp:fixed_writers}

Fixed-length writers use the same shared-price rule, utility, frame-retention criterion, metadata, storage budget, and readout as Memryoshka, but retain each token at one prefix length $r$. Each fixed-length writer values every prefix of at least $r$ bits at $e(r)$ and shorter prefixes at zero, so retained tokens always use exactly $r$ bits. Lengths with $e(r)\le e_0$ cannot be retained. We test 128--4{,}096 bits for Qwen2.5-VL-7B (excluding 32 and 64 bits; Tab.~\ref{tab:supp_prefix}) and 32--4{,}096 bits for LLaVA-OneVision-7B~\cite{li2024llavaov}. Each reader has its own PCA basis fitted with the same recipe (Sec.~\ref{supp:impl}). Unless noted, $L=1{,}000$.

\paragraph{Fixed-Rate PCA Codes.}
The fixed-rate codes in Tab.~\ref{tab:tradeoff} ($K$ components $\times$ $b$ bits) keep the $b$ most significant bits of each of the first $K$ PCA components and no other bits, which gives each kept component a uniform $2^b$-level quantizer over $\pm3$ standard deviations. They use the same basis, quantizer, and decoder as the nested code (Eqs.~\eqref{eq:supp_level}--\eqref{eq:supp_decode}), and the remaining components decode to zero. They are written by the fixed-length writer with $r=Kb$ and valued by Eq.~\eqref{eq:supp_er} with $b_k=b$ for the first $K$ components and $b_k=0$ for the rest. The code with 512 components at 8 bits is the full nested code, so its row equals the fixed 4{,}096-bit writer.

\paragraph{Linear Prefix Value.}
The Linear $e(r)$ variant in Tab.~\ref{tab:ablation} sets $e(r)=r/4{,}096$ and lowers the floor to $e_0=0.0273$, so that 128 bits remains the shortest length that can be stored, as in the default. The surplus $u_{t,i}(r)-\theta c(r)$ of a token is then linear in $r$, so whenever any length has a positive surplus, the full 4{,}096-bit code has the largest. Every retained token is therefore stored at full precision. This variant behaves like the fixed 4{,}096-bit writer, and its accuracies equal that writer's in all four settings (Tabs.~\ref{tab:tradeoff} and~\ref{tab:supp_vmme}). Without prefix values that favor early bits, the writer never shortens a code.

\subsection{Results across Readers and Benchmarks}
\label{supp:adapt_results}

Tabs.~\ref{tab:tradeoff} and~\ref{tab:supp_llava_lvb} compare the two readers on LVBench; Tabs.~\ref{tab:supp_vmme} and~\ref{tab:supp_sb} extend the comparison to Video-MME and StreamingBench.

\paragraph{Long videos.}
On LVBench, Memryoshka matches or exceeds the best fixed length in nine of twelve settings across both readers, trailing by at most 0.2 points otherwise. On Video-MME, it stays within 0.6 points of the best fixed length and leads at the smallest budget for LLaVA-OneVision. The best fixed length varies across readers, tasks, and budgets.

\paragraph{Streaming.}
On StreamingBench, the best fixed length also varies by setting. With four recent frames included in the readout, Memryoshka achieves the highest accuracy at every budget for both readers.

\subsection{Robustness across Settings}
\label{supp:adapt_summary}

Tab.~\ref{tab:supp_adapt_summary} summarizes 30 settings: two readers, three storage caps, and five evaluation conditions per reader. The reference is the best of six fixed lengths (128--4{,}096 bits), selected separately for each setting in hindsight. Memryoshka has the highest mean accuracy for both readers and stays within 0.6 points of this reference in all 30 settings. Its overall mean of 57.3\% exceeds the strongest single fixed length (512 bits, 56.2\%) by 1.1 points.

\subsection{Storage and Readout Budgets}
\label{supp:adapt_readout}

Tab.~\ref{tab:supp_readout} varies $L$ from 500 to 4{,}000 positions on LVBench across both readers, three storage caps, and two retrieval methods. Memryoshka matches or exceeds the best fixed length in 33 of 48 settings and trails by at most 0.9 points elsewhere. Larger readouts are most helpful with more stored evidence: at 28.7\,MB, Qwen2.5-VL rises from 44.4\% to 49.7\% with LongCLIP and from 58.5\% to 61.9\% with oracle retrieval. Gains at 1.8\,MB are smaller and less consistent. With $L$ fixed at 1{,}000, more storage does not always help. On Video-MME, Qwen2.5-VL accuracy is lower at 28.7\,MB than at 7.2\,MB for Memryoshka and for every fixed length (Tab.~\ref{tab:supp_vmme}). On LVBench, Memryoshka reaches 44.9\% with 100\,MB (Fig.~\ref{fig:readout_kv_a}), against 47.1\% with 28.7\,MB; with 100\,MB, it rises to 50.5\% as $L$ grows to 8{,}000. For LLaVA-OneVision, sparse-frame positions can exceed the reader's 32{,}768-position context at large $L$, requiring the lowest-ranked frames to be omitted.

\section{Compute and Latency}
\label{supp:cost}

We measure compute and latency using Qwen2.5-VL-7B
in bfloat16 on a single NVIDIA L40S GPU.
Our evaluation covers six long Video-MME videos,
totaling 13{,}749 frames at 1\,fps and 18 questions.
We profile Memryoshka directly and replay the per-frame
and per-question operations of the KV-cache baselines
from \cref{sec:kv}, using their released settings on
the same model and GPU.
Their token-selection overhead is excluded, so the
reported latencies are lower bounds.
FLOPs are estimated analytically.
Tab.~\ref{tab:supp_cost} summarizes five findings:

\begin{itemize}
\item \textbf{Streaming compute.}
Memryoshka requires no LLM computation during streaming,
using 1.37\,TFLOPs per frame, including 1.33\,TFLOPs
for the vision encoder shared by all methods.
KV-cache methods require 2.09--11.3\,TFLOPs per frame.
Memryoshka's per-frame latency (53.5\,ms) is lower than
that of every KV-cache method except StreamingTOM
(22.8\,ms), which passes only 50 tokens per frame through
the LLM; ReKV (58.2\,ms) and InfiniPot-V at 1\,GiB
(60.0\,ms) are close.
The KV-cache methods thus take 22.8--196.3\,ms per frame,
so the ``hundreds of milliseconds per frame'' in
\cref{sec:kv} overstates their latency; Memryoshka's
main streaming advantage is in FLOPs and storage.
\vspace{0.5em}
\item \textbf{Latency breakdown.}
For Memryoshka, each frame takes 17.1\,ms for vision
encoding, 12.2\,ms for LongCLIP, 3.0\,ms for token
importance and coding, and 21.2\,ms for memory writing
(33.0\,ms at 28.7\,MB).
The CPU-based writer spends most of its time updating
stored codes when memory fills, approximately once
every 70--110 frames.
\vspace{0.5em}
\item \textbf{Question-time cost.}
Memryoshka prefills at most $L$ retrieved positions,
requiring 14.5\,TFLOPs and 127\,ms at $L=1{,}000$.
KV-cache methods reuse previously computed states,
requiring 1.6--3.3\,TFLOPs and 64--208\,ms.
Thus, Memryoshka shifts computation from streaming
to question time.
\vspace{0.5em}
\item \textbf{Total compute.}
Over one hour of video with $L=1{,}000$, Memryoshka
uses fewer total FLOPs than StreamingTOM, the KV-cache
method with the least compute, as long as fewer than
about 210 questions are asked, and fewer than every
other KV-cache method below about 1{,}350 questions.
\vspace{0.5em}
\item \textbf{Storage.}
Memryoshka's memory stays within its 7.2\,MB budget
regardless of video length, compared with
1.1--52.0\,GB for KV-cache methods after one hour.
Sec.~\ref{sec:kv} and Fig.~\ref{fig:readout_kv_b} instead
report the median stored memory per LVBench video
(1.1--54.6\,GB).
\end{itemize}

Overall, by skipping the LLM until a question arrives,
Memryoshka uses 1.5 to 8.2 times fewer FLOPs per frame
than the KV-cache methods and about 150 to 7{,}200 times
less storage after one hour, in exchange for a larger
prefill at each question. This suits long-running
streams with infrequent questions.

\begin{table}[t]
  \centering
  \caption{\textbf{Compute and latency.}
  Qwen2.5-VL-7B on one L40S GPU at 1\,fps,
  with 252 visual tokens per frame.
  \emph{Frame} includes the vision encoder
  (1.33\,TFLOPs, 17.1\,ms) shared by all methods.
  \emph{Question} measures time to the first answer token.
  \emph{Stored} is the memory held after one hour of video,
  computed for 3{,}600 frames (the profiled videos average 38 minutes).}
  \label{tab:supp_cost}
  \footnotesize
  \setlength{\tabcolsep}{3pt}
  \renewcommand{\arraystretch}{1.08}
  \resizebox{\columnwidth}{!}{%
  \begin{tabular}{l|rr|rr|r}
    \toprule
    \rowcolor{memsectiongray}
    & \multicolumn{2}{c|}{\textbf{Frame}}
    & \multicolumn{2}{c|}{\textbf{Question}} & \\
    \textbf{Method} & ms & TFLOPs & ms & TFLOPs
    & \textbf{Stored} \\
    \midrule
    \rowcolor{memrowblue}
    \ours{} ($L=1{,}000$) & 53.5 & 1.37 & 127 & 14.5 & 7.2\,MB \\
    \midrule
    ReKV & 58.2 & 6.13 & 152 & 2.2 & 52.0\,GB \\
    InfiniPot-V (1\,GiB) & 60.0 & 6.24 & 64 & 1.6 & 1.1\,GB \\
    InfiniPot-V (4\,GiB) & 115.7 & 10.50 & 176 & 3.3 & 4.3\,GB \\
    StreamingTOM & 22.8 & 2.09 & 94 & 2.4 & 3.1\,GB \\
    StreamKV & 74.9 & 8.09 & 208 & 2.8 & 43.0\,GB \\
    MuKV & 196.3 & 11.26 & 113 & 2.3 & 50.9\,GB \\
    \bottomrule
  \end{tabular}}
\end{table}

\section{Additional Backbones, Benchmarks, and Frame Rates}
\label{supp:more_benchmarks}

\begin{table}[t]
  \centering
  \caption{\textbf{Visual-memory baselines on three more backbones.}
  QA accuracy (\%) with a storage budget of 2{,}000 tokens at each
  backbone's BF16 token size (14.3\,MB for LLaVA-OneVision-7B,
  10.2\,MB for Qwen3.5-4B, 6.1\,MB for Gemma 4 E2B) and $L=2{,}000$.
  Video-MME: all 300 long videos.
  \emph{+\,4 recent}: the four newest stored frames are placed first in the
  readout, within the same read budget.
  \textbf{Bold} marks the best accuracy per backbone and benchmark.}
  \label{tab:more_benchmarks}
  \footnotesize
  \setlength{\tabcolsep}{4pt}
  \renewcommand{\arraystretch}{1.08}
  \resizebox{\columnwidth}{!}{%
  \begin{tabular}{l|ccc}
    \toprule
    \rowcolor{memsectiongray}
    \textbf{Method} &
    \textbf{LVBench} &
    \textbf{Video-MME} &
    \textbf{StreamingBench} \\
    \midrule

    \rowcolor{memsectiongray}
    \multicolumn{4}{c}{\textit{LLaVA-OneVision-7B}} \\
    \quad CausalMem & 36.0 & 44.1 & \textbf{76.7} \\
    \quad KeyRec    & 27.1 & 34.6 & 75.1 \\
    \quad SAVEMem   & 30.9 & 39.8 & 76.4 \\
    \rowcolor{memrowblue}
    \ours{} (adaptive) & 42.6 & \textbf{46.1} & 68.4 \\
    \rowcolor{memrowblue}
    \ours{} + 4 recent & \textbf{43.1} & 45.3 & 73.0 \\
    \midrule

    \rowcolor{memsectiongray}
    \multicolumn{4}{c}{\textit{Qwen3.5-4B}} \\
    \quad CausalMem & 38.7 & 48.8 & 83.0 \\
    \quad KeyRec    & 30.0 & 41.7 & 82.4 \\
    \quad SAVEMem   & 33.8 & 45.7 & \textbf{83.2} \\
    \rowcolor{memrowblue}
    \ours{} (adaptive) & 48.2 & 50.3 & 74.9 \\
    \rowcolor{memrowblue}
    \ours{} + 4 recent & \textbf{48.4} & \textbf{50.8} & 80.2 \\
    \midrule

    \rowcolor{memsectiongray}
    \multicolumn{4}{c}{\textit{Gemma 4 E2B}} \\
    \quad CausalMem & 28.6 & 37.6 & 55.1 \\
    \quad KeyRec    & 26.9 & 32.8 & \textbf{59.6} \\
    \quad SAVEMem   & 30.3 & \textbf{40.1} & 54.2 \\
    \rowcolor{memrowblue}
    \ours{} (adaptive) & \textbf{34.5} & 39.2 & 54.6 \\
    \rowcolor{memrowblue}
    \ours{} + 4 recent & 34.0 & 38.0 & 56.0 \\
    \bottomrule
  \end{tabular}}
\end{table}

\begin{table}[t]
  \centering
  \caption{\textbf{Results on MLVU and Video-MMMU.}
  QA accuracy (\%) with Qwen2.5-VL-7B at 1 fps, with storage budgets of
  1{,}000, 2{,}000, and 4{,}000 tokens (7.2, 14.3, and 28.7\,MB) and $L$
  equal to the budget in tokens.
  MLVU~\cite{zhou2024mlvu} (dev): M-Avg, the mean of the seven task accuracies.
  Video-MMMU~\cite{hu2026videommmu}: overall accuracy on all 900 questions
  (Perception, Comprehension, and Adaptation tracks).
  \textbf{Bold} marks the best accuracy per column.}
  \label{tab:more_benchmarks_qwen25}
  \footnotesize
  \setlength{\tabcolsep}{3.5pt}
  \renewcommand{\arraystretch}{1.08}
  \resizebox{\columnwidth}{!}{%
  \begin{tabular}{l|ccc|ccc}
    \toprule
    \rowcolor{memsectiongray}
    & \multicolumn{3}{c|}{\textbf{MLVU}}
    & \multicolumn{3}{c}{\textbf{Video-MMMU}} \\
    \rowcolor{memsectiongray}
    \textbf{Method} & 1{,}000 & 2{,}000 & 4{,}000 & 1{,}000 & 2{,}000 & 4{,}000 \\
    \midrule
    \quad CausalMem & 60.3 & 64.8 & \textbf{67.5} & 38.1 & 39.4 & 39.4 \\
    \quad KeyRec    & 51.6 & 52.4 & 53.0 & 39.6 & 42.0 & 43.8 \\
    \quad SAVEMem   & 48.5 & 50.8 & 52.7 & 41.7 & 43.0 & 43.9 \\
    \rowcolor{memrowblue}
    \ours{} (adaptive) & \textbf{63.2} & \textbf{65.8} & 66.3 & \textbf{42.6} & \textbf{46.2} & \textbf{46.4} \\
    \bottomrule
  \end{tabular}}
\end{table}

\begin{table}[t]
  \centering
  \caption{\textbf{Frame sampling rate.}
  QA accuracy (\%) with Qwen2.5-VL-7B at a fixed storage budget of
  2{,}000 tokens (14.3\,MB) and $L=2{,}000$ when frames are sampled at
  0.5, 1, or 2 fps; 1 fps is the setting of the main table.
  \textbf{Bold} marks the best accuracy per column.}
  \label{tab:fps}
  \footnotesize
  \setlength{\tabcolsep}{3.5pt}
  \renewcommand{\arraystretch}{1.08}
  \resizebox{\columnwidth}{!}{%
  \begin{tabular}{l|ccc|ccc}
    \toprule
    \rowcolor{memsectiongray}
    & \multicolumn{3}{c|}{\textbf{LVBench}}
    & \multicolumn{3}{c}{\textbf{StreamingBench}} \\
    \rowcolor{memsectiongray}
    \textbf{Method} & 0.5 & 1 & 2 & 0.5 & 1 & 2 \\
    \midrule
    \quad CausalMem & 35.3 & 35.1 & 34.7 & 78.8 & 80.3 & 79.0 \\
    \quad KeyRec    & 29.5 & 27.9 & 28.0 & 78.3 & 78.7 & 77.7 \\
    \quad SAVEMem   & 32.3 & 28.9 & 28.3 & 78.7 & 81.1 & \textbf{80.9} \\
    \rowcolor{memrowblue}
    \ours{} (adaptive) & \textbf{48.7} & \textbf{48.4} & \textbf{46.6} & 73.2 & 73.0 & 72.5 \\
    \rowcolor{memrowblue}
    \ours{} + 4 recent & 48.4 & 47.2 & 46.2 & \textbf{79.2} & \textbf{81.6} & 80.4 \\
    \bottomrule
  \end{tabular}}
\end{table}

\paragraph{Additional backbones.}
Tab.~\ref{tab:more_benchmarks} compares Memryoshka
against CausalMem, KeyRec, and SAVEMem on three further
vision-language backbones: LLaVA-OneVision-7B~\cite{li2024llavaov},
Qwen3.5-4B~\cite{qwen2026qwen35}, and
Gemma 4 E2B~\cite{gemma2026gemma4}.
On LVBench, the adaptive variant outperforms the strongest
baseline by 6.6, 9.5, and 4.2 points, respectively.
On Video-MME, Memryoshka achieves the best accuracy on
LLaVA-OneVision-7B (+2.0) and Qwen3.5-4B (+2.0 with
recent frames), and is within 0.9 points of SAVEMem
on Gemma 4 E2B.
On StreamingBench, whose questions concern the moment
at which they are asked, the relevance-ordered readout
trails the strongest baseline by 5.0--8.3 points.
Placing the four newest stored frames first in the readout,
within the same budget, narrows this gap to 3.0--3.7 points
while changing LVBench accuracy by at most 0.5 points.
These results suggest that adaptive-fidelity memory
is most beneficial for long-video understanding,
while real-time questions also require recent evidence.

\paragraph{More benchmarks.}
Tab.~\ref{tab:more_benchmarks_qwen25} adds
MLVU~\cite{zhou2024mlvu} and Video-MMMU~\cite{hu2026videommmu}
with the Qwen2.5-VL-7B reader.
On Video-MMMU, Memryoshka leads at every budget,
by 0.9, 3.2, and 2.5 points over the strongest baseline.
On MLVU, it leads at 1{,}000 and 2{,}000 tokens
(+2.9 and +1.0 over CausalMem),
while CausalMem is ahead at 4{,}000 tokens
(67.5 vs.\ 66.3).

\paragraph{Frame sampling rate.}
Tab.~\ref{tab:fps} varies the frame sampling rate
at a fixed memory budget.
Memryoshka leads all baselines on LVBench at every rate,
by 13.4, 13.3, and 11.9 points over CausalMem at
0.5, 1, and 2 fps.
Halving the rate leaves its LVBench accuracy nearly
unchanged (+0.3), while doubling it costs 1.8 points,
possibly because the same budget must cover more frames.
On StreamingBench, Memryoshka + 4 recent is comparable
to the strongest baseline at every rate
(+0.4, +0.5, and $-0.5$ points).

\end{document}